\documentclass[letterpaper]{article} 
\usepackage{aaai2027}  
\usepackage[hyphens]{url}  
\usepackage{graphicx} 
\usepackage{natbib}  
\usepackage{caption} 
\usepackage{algorithm}
\usepackage{algorithmic}

\usepackage{newfloat}
\usepackage{booktabs}
\usepackage{booktabs}
\usepackage{listings}
\DeclareCaptionStyle{ruled}{labelfont=normalfont,labelsep=colon,strut=off} 
\floatstyle{ruled}
\newfloat{listing}{tb}{lst}{}
\floatname{listing}{Listing}

\usepackage{booktabs}
\usepackage{subcaption}
\usepackage{multirow}
\usepackage{amssymb}

\usepackage{amsmath}

\title{GPEvac: GNN-Based PPO for Adaptive Evacuation Routing During Shooting Events}
\author {
    Daniel Perkins\textsuperscript{\rm 1,\rm 2},
    Subhadeep Chakraborty\textsuperscript{\rm 2}
}
\affiliations {
    \textsuperscript{\rm 1}The Bredesen Center for Interdisciplinary Research and Graduate Education\\
    \textsuperscript{\rm 2}Department of Mechanical and Aerospace Engineering, University of Tennessee\\
    dperki16@vols.utk.edu, schakrab@utk.edu
}

\begin{document}

\maketitle

\begin{abstract}
The sharp increase in mass shootings underscores an urgent need for systems that guide victims to safety in real time. An effective evacuation system must minimize threat exposure while also accounting for adversarial uncertainty and crowding dynamics. Current methods in the literature are rigidly constrained to layout-specific policies and computationally intractable in large-scale layouts, while practical guidelines simply advise victims to ``run'', ``hide'', or ``fight''. We propose GPEvac: a GNN-based PPO framework that computes adaptive evacuation routes during shooting events. To capture both local and long-distance dependencies, we introduce an edge-first sequential message-passing scheme with a learnable virtual global node. The resulting graph embeddings are integrated into a permutation-invariant scoring mechanism that allows a single learned policy to operate across building layouts of diverse topologies and sizes. Through extensive simulation, we show that GPEvac outperforms intelligent baselines across distinct architectural layouts, significantly reducing total threat exposure. Crucially, the system computes global evacuation routes in just 14.73 ms on local CPU hardware, enabling seamless integration with live surveillance systems. In addition to saving lives during shooting events, the methodologies developed are transferable to other graph-structured decision-making domains, including critical infrastructure, intelligent transportation systems, and adaptive sensor networks.
\end{abstract}


\section{Introduction}
\label{sec:intro}
The Gun Violence Archive \cite{klein2025gunviolencearchive} indicates that between 2020 and 2024 there were 3,106 mass shootings in America, 1,295 more than the previous five years. This sharp increase in violence has led to the death of 12,617 children and teenagers in the last ten years, underscoring that shootings in large public spaces are a significant threat to public safety~\cite{fbi_active_shooter_2014}. While law enforcement has focused on prevention and rapid response, little research has been done to address how evacuees should act during the critical period before a threat is contained. 

The U.S. Department of Homeland Security currently recommends the ``Run–Hide–Fight'' protocol for active shooter situations in schools and workplaces. This framework requires individuals to make rapid situational assessments: evacuating if a viable escape route exists, securing a safe hiding location, or physically confronting the attacker as a last resort~\cite{DHSActiveShooterPreparedness}. While existing agent-based simulations have evaluated this protocol in isolated scenarios~\cite{Lee2019RunHideFightABM}, the ``Run–Hide–Fight” strategy ultimately lacks the adaptability required for diverse environments~\cite{Rummaneethorn2024RunHideFightHospital}. Furthermore, relying on decentralized, uninformed decision-making often leads to suboptimal evacuation routes, congestion, and inadvertent encounters with the active threat~\cite{akbarzadeh2025visitor}. 

The ``Run-Hide-Fight'' protocol fails to equip evacuees with an optimized escape plan tailored to the building layouts and dynamic conditions of an active-shooter situation. We propose an evacuation routing system that computes optimal escape paths that minimize both overall evacuation time and the duration evacuees are exposed to the threat. By employing computer vision techniques~\cite{redmon2016yolo, maggiolino2023deepocs, robinson2026rfdetr, Waite2023ActiveSD} to continuously track the assailant and measure crowd density, our model leverages a global view of the environment. This real-time information enables the calculation of complex, adaptive routes that systematically avoid the dynamic threat while mitigating severe congestion bottlenecks.

\begin{figure}[h]
    \centering
    \includegraphics[width=\columnwidth]{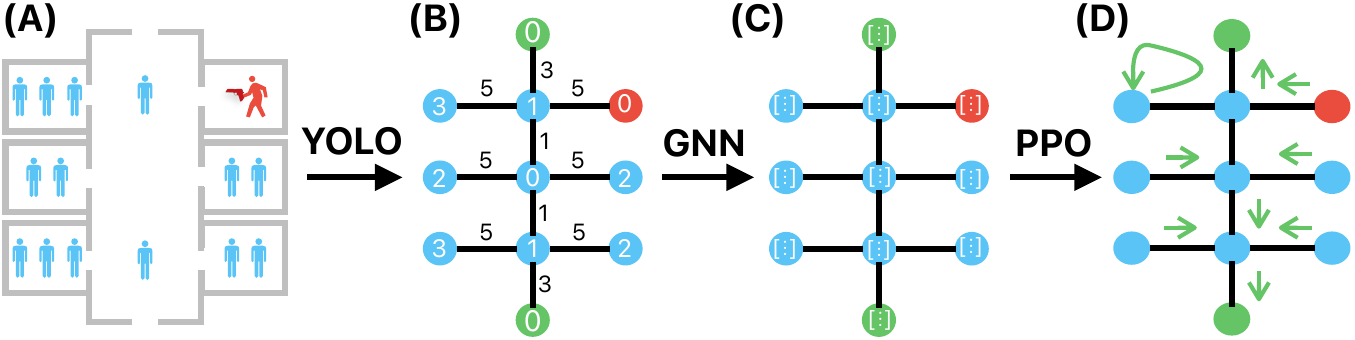}
    \caption{The high-level pipeline of GPEvac. (A) An example building with an active shooter (red) and evacuees (blue). (B) Simplified graphical representation with the number of evacuees as the node features and travel time as the edge features. (C) GNN-generated node embeddings. (D) PPO Output: each green arrow denotes the target destination.}
    \label{fig:ppo_pipeline}
\end{figure}

While advances in deep reinforcement learning enable adaptive control in dynamic environments, their potential for real-time evacuation routing remains underexplored. We propose a model that represents building layouts as graphs and employs an edge-centric graph neural network with a virtual global node. This specialized network produces context-aware node embeddings that encode spatial relationships, congestion risks, and dynamic threat proximity. These embeddings then drive a Proximal Policy Optimization (PPO) framework, where a permutation-invariant scoring mechanism allows the network to optimize evacuation routes across varying graph topologies (Figure \ref{fig:ppo_pipeline}). Ultimately, this approach enables a single neural network to be utilized across multiple different building layouts, providing a tool that can actively minimize threat exposure in diverse scenarios.

\section{Related Works}
\label{sec:related_works}

Traditional pathfinding algorithms, such as Dijkstra’s algorithm~\cite{dijkstra1959} and A*~\cite{hart1968formal} can compute optimal escape routes in static environments. However, modeling the shooter’s movement and accounting for crowd congestion introduces substantial computational complexity, rendering these methods infeasible for real-time decision support in large-scale scenarios.

\citeauthor{Gunn2017} pioneered approaches to this problem, using a stochastic dynamic programming approach to optimize evacuation routing~\cite{Gunn2017}. ASTERS advanced this research by modeling building layouts as graphs and applying a non-homogeneous semi-Markov decision process~\cite{LavalleRivera2023}. This work was later extended to include capacity constraints~\cite{LavalleRivera2025}. While these models demonstrate strong performance in controlled environments, they remain computationally intractable for large-scale scenarios and are rigidly constrained to specific layouts. Scalable models for optimal escape routing in complex, unfamiliar environments demand modern deep learning paradigms.



\subsection{Deep Reinforcement Learning (DRL)}
\label{sec:drl}


DRL has emerged as a powerful framework for adaptive decision-making in complex environments. Value-based methods \cite{Mnih2013} work well in discrete domains but struggle to scale in high-dimensional action spaces. Actor-critic networks address this by training two separate neural networks simultaneously: a critic to evaluate states, and an actor to predict the optimal action~\cite{Mnih2016}. Proximal Policy Optimization (PPO) is a widely adopted actor-critic approach, enhancing training stability via a clipped surrogate objective~\cite{Schulman2017}. 

DRL has been adopted for evacuation routing. However, existing frameworks struggle to balance topological abstraction with dynamic threat adaptability. Microscopic continuous-space models~\cite{xu2020simulating, Zhang2021DeepReinforcement, Huang2023DynamicScanning} and hierarchical leader-follower approaches~\cite{zhang2024double} successfully optimize localized obstacle avoidance. But they lack the graph-based abstraction required for complex building layouts. Graph-based methods like EvacuAI~\cite{Rosa2023EvacuAI} and ReinforceRouting~\cite{li2023reinforcement} model spatial structure. However, they either discard topology or rely on static node2vec embeddings, lacking the dynamic message-passing required to aggregate real-time state changes.


\subsection{Graph Neural Networks (GNNs)}
\label{sec:gnns}

GNNs provide a natural framework for pathfinding tasks~\cite{Liu2025}. By replacing fully connected layers with message passing, these networks explicitly encode topological structure, producing permutation-invariant node embeddings that scale seamlessly across varying graph sizes and structures~\cite{Kipf2017, gilmer2017neural}. 

To maximize expressive power, various architectures employ distinct aggregation strategies. Graph Isomorphism Networks (GINEs)~\cite{xu2018how} utilize sum aggregation, and Principal Neighbourhood Aggregation (PNA)~\cite{corso2020principal} combines multiple aggregators (mean, max, min, standard deviation) with degree-based scaling. For dynamic information routing, Residual Gated GCNs (ResGated)~\cite{Bresson2017} apply edge gating, Graph Attention Networks (GATs)~\cite{Velickovic2018} use local softmax attention, and Graph Transformers~\cite{yun2019graph} leverage global attention. We systematically compare each variant in our ablation studies to identify the optimal architecture.

A fundamental limitation of GNNs is that they only expand receptive fields by one hop per layer, failing to capture the global information in large-diameter graphs. While stacking layers expands this field, it induces oversmoothing, causing node embeddings to collapse~\cite{chen2020measuring}. To mitigate these bottlenecks, our framework incorporates a virtual global node~\cite{gilmer2017neural, battaglia2018relational, hu2020open, Medina2026GNNPolymers} to efficiently propagate long-range dependencies and generate the context-aware embeddings required for optimal evacuation policies.


\subsection{GNN-Based Deep Reinforcement Learning}
\label{sec:graph-based-drl}

Integrating GNNs with DRL enables policies to generalize across unseen topologies~\cite{almasan2019deep}. To handle distributed problems, these frameworks are often extended to multi-agent settings~\cite{Jaderberg2019, Yu2022}. For instance, decentralized graph-based MADRL successfully optimizes resource allocation~\cite{Shao2021} and traffic control~\cite{Zeng2022}. However, assigning individual rewards and critics inherently limits coordination. 

Centralized methods unify the reward, pooling the GNN-extracted embeddings to inform a centralized critic~\cite{Kotecha2025, Zhao2025}. While these methods improve coordination, real-world implementations frequently face issues with non-stationarity, instability, and high sample complexity~\cite{Papoudakis2019DealingWN, Sharma2021SurveyMARL}. Because these multi-agent bottlenecks critically restrict deployment in safety-critical scenarios, our framework bypasses them entirely by abstracting the multi-node routing problem into a robust single-agent formulation.

More closely related to our domain, a few works have begun applying GNN-based RL to evacuation routing. GreyGNN-MARL~\cite{Zhang2025} uses GNNs and Q-learning for wildfires, but its iterative meta-heuristic limits real-time scalability. DRAGON integrates GCNs with Q-learning for indoor evacuation, but optimizes single paths rather than coordinated global evacuations~\cite{Abouelaziz2026DRAGON}. \citeauthor{wu2025toward} apply GNNs and PPO for evacuation in large buildings~\cite{wu2025toward}. However, without a permutation-invariant actor, their architecture cannot deploy a single policy across diverse topologies. Furthermore, all these existing frameworks assume predictable hazard dynamics, leaving them ill-equipped to handle the highly stochastic nature of an active shooter.

\section{Methodology}
\label{sec:methodology}

We propose GPEvac (GNN-PPO Evacuation), an evacuation routing algorithm that combines GNNs with PPO to navigate individuals safely away from active threats. The pipeline begins by extracting real-time data using prior work to track the shooter and count evacuees~\cite{Waite2023ActiveSD}. This real-time observational data is structured into a dynamic graph. A hand-crafted GNN then processes this graph to generate rich node embeddings, which serve as the state representation for the PPO agent to formulate safe routing decisions for each room. The complete pipeline is summarized in Figure \ref{fig:ppo_pipeline}.

\subsection{Simulation Environment}
\label{sec:sim_env}

\begin{figure}[h]
\centering
    \begin{minipage}[c]{0.48\textwidth}
    \centering
    {%
      \includegraphics[width=0.55\textwidth]{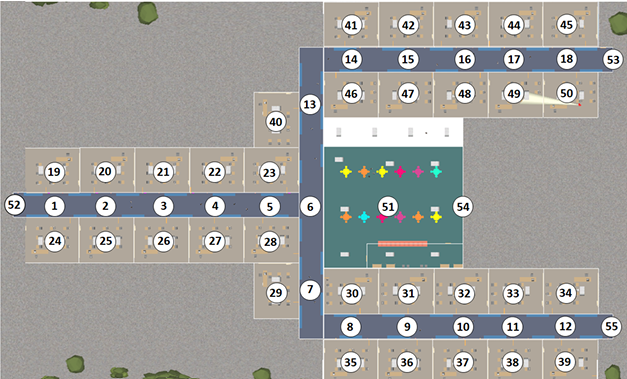}%
    }
    \end{minipage}%
    \hfill
    \begin{minipage}[c]{0.48\textwidth}
    \centering
    {%
      \includegraphics[width=0.85\textwidth]{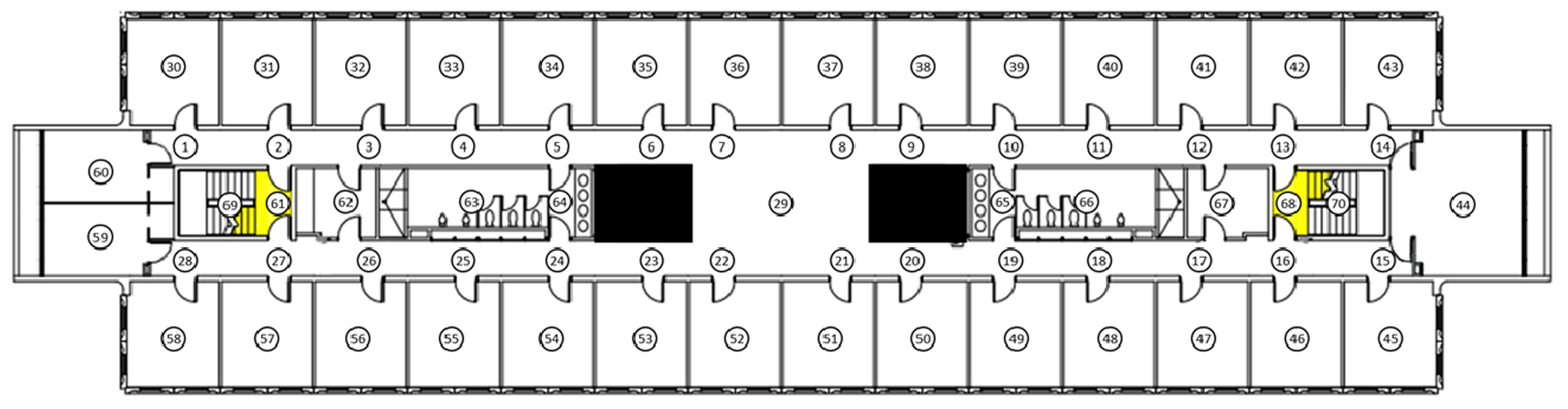}%
    }
    \end{minipage}%
\caption{Floor plans of the Acyclic (top) and Cyclic (bottom) school layouts~\cite{LavalleRivera2023,LavalleRivera2025}.}
\label{fig:school-layouts}
\end{figure}

Because video footage is limited and live experiments are ethically impossible, we develop a simulation environment that synthesizes realistic shooter and evacuee dynamics. For practical applicability and reduced overhead, the movement is modeled as a dynamic graph that replicates the expected output of real-time tracking algorithms~\cite{Waite2023ActiveSD}. To ensure robustness to fundamentally different topologies, we train a single model across two distinct building layouts: one acyclic layout with strict structural bottlenecks, and one cyclic layout with multiple feasible escape routes (Figure \ref{fig:school-layouts}).

\begin{figure}[h]
    \centering
    \includegraphics[width=0.9\columnwidth]{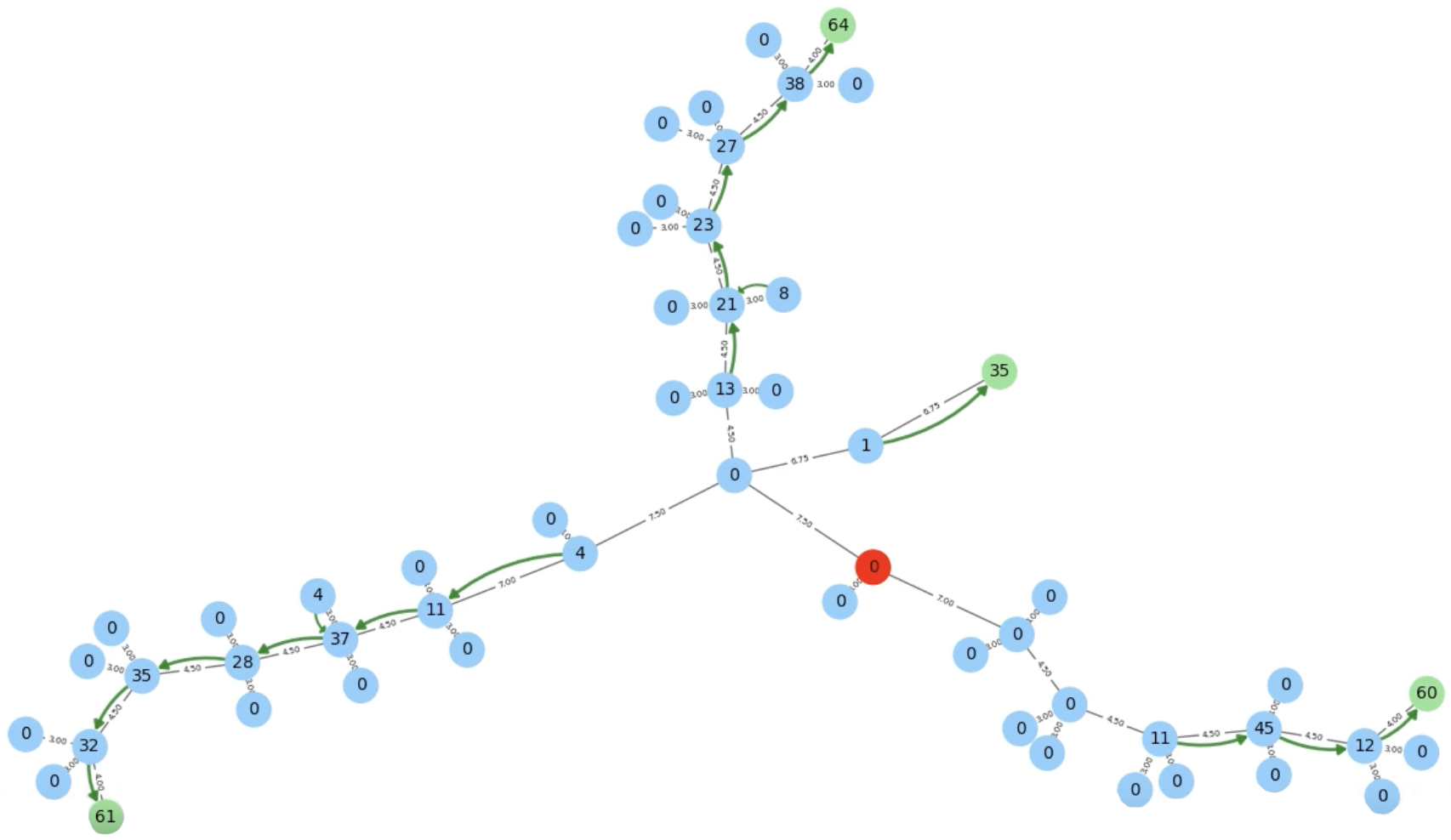}
    \caption{Graphical representation of the acyclic school in Figure \ref{fig:school-layouts}. Each node is a room with $n$ people. The green nodes define the exits and the red node the location of the threat. The arrows are an example output of the GPEvac.}
    \label{fig:example_graph}
\end{figure}

Each building is modeled as a graph $G=(V,E)$, where $V=\{1, \dots, N\}$ and $E \subseteq V \times V$ are the nodes and edges (Figure \ref{fig:example_graph}). Nodes correspond to rooms, with hallways discretized into multiple nodes based on room adjacency. Edges represent the connections between rooms. To allow evacuees to stay in place, we include self-loops $(i,i) \in E$.

To represent the environment state, each node $i \in V$ is encoded as a raw feature vector $v_i \in \mathbb{R}^M$. As detailed in the Supplementary Material, these vectors capture dynamic variables, such as room occupancy and the presence of a threat. They also include structural properties, such as centrality metrics~\cite{freeman1978centrality} and graph distances to the exits and threat(s). Similarly, each edge $(i,j) \in E$ is represented by a feature vector $e_{i,j} \in \mathbb{R}^L$ that encodes relational attributes, including traversal time and the occupancy gradient. 

For each initialized environment, occupants are randomly distributed to mirror realistic school building patterns while ensuring episode variability. Hallway occupancies (ranging from $0$ to $4$ people) are drawn from $\text{Categorical}(0.85, 0.10, 0.03, 0.015, 0.005)$. $80\%$ of classrooms are occupied, with occupant counts sampled from $\mathcal{N}(18.3, 9)$ to closely align with national public school averages~\cite{nces_ntps_2020_21_classsize}. Finally, active shooters are randomly placed on one or more nodes.

Movement through the environment is modeled as a discretized flow between nodes. For each node, the routing algorithm selects a neighbor to receive the next group of evacuees. The volume of individuals moved during this action is dictated by the corresponding edge weight, where smaller weights accommodate larger groups. Because physical transit is not instantaneous, this transition is distributed over 5 discrete time steps, incrementally updating the respective node ($v_i, v_j$) and edge features ($e_{i,j}, e_{j,i}$) at each step.

The movement of the active shooter follows a similar process, but its routing logic is intentionally stochastic. In real-world scenarios, an assailant's intent is highly unpredictable. Furthermore, operating under incorrect deterministic assumptions carries a severe penalty. To account for this uncertainty, we model the threat's movement as a random walk, uniformly sampling the next target from the set of all adjacent nodes. This prevents the model from overfitting to specific behavioral profiles, ensuring the learned evacuation policies remain robust to unpredictable threat dynamics.

\subsubsection{Reward Function}
\label{sec:reward_function}

To minimize both contact with the shooter and overall evacuation time, we design a novel reward function (Equation \ref{eq_r_t}). Let $n_i$ denote the occupant count in node $i$, with $n_{\text{total}}$, $n_{\text{remaining}}$, and $n_{\text{exit}}$ representing the initial, current, and newly escaped populations. Additionally, let $d_i$ be the distance from node $i$ to the threat. Define $\Omega=1$ if all people reach an exit, $\Omega=-1$ if the episode terminates without everyone escaping, and $\Omega=0$ if the episode has not yet terminated. Let $R_{\text{rate}}$, $R_{\text{zero}}$, $R_{\text{step}}$, $R_{\text{threat}}$, $R_{\text{exit}}$, $R_{\text{finished}}$, and $R_{\text{time}}$ be user-defined hyperparameters.
\begin{align}
    \beta_{i} &= \max\left(0, \frac{R_{\text{rate}}^{d_{i}/R_{\text{step}}}-R_{\text{rate}}^{R_{\text{zero}}/R_{\text{step}}}}{1-R_{\text{rate}}^{R_{\text{zero}}/R_{\text{step}}}}\right)  \label{eq:beta_ij} \\
    r_{\text{threat}} &= R_{\text{threat}}\sum_i\beta_{i}\left(\frac{n_i}{n_{\text{total}}}\right) \label{eq_r_threat} \\
    r_{\text{escape}} &=  R_{\text{exit}}\left(\frac{n_{\text{exit}}}{n_{\text{total}}}\right) + R_{\text{finished}}\left(\Omega\right)\label{eq:r_escape} \\
    r_{\text{time}} &= R_{\text{time}}\left(0.1+\frac{n_{\text{remaining}}}{n_{\text{total}}}\right) \label{eq:r_time_penalty} \\
    r &= -r_{\text{threat}}+r_{\text{escape}}- r_{\text{time}} \label{eq_r_t}
\end{align}

The most important part of the reward function is $r_{\text{threat}}$ (Equation \ref{eq_r_threat}), which penalizes the model for each evacuee close to the threat. The local penalty $\beta_{i}$ is $1$ when $d_{i}=0$, decays exponentially by a factor of $R_{\text{rate}}$ for every increase in distance of  $R_{\text{step}}$, and is $0$ whenever $d_{i}\geq R_{\text{zero}}$. This induces a sharp penalty when people are in the same room as the threat and a small penalty when they are far away.


To ensure that evacuees actively move toward safety rather than merely hiding, we supplement $r_{\text{threat}}$, using evacuation as a proxy for overall risk reduction. $r_{\text{escape}}$ rewards the model for every individual who successfully evacuates, alongside a terminal bonus if the entire population escapes (Equation \ref{eq:r_escape}). Additionally, $r_{\text{time}}$ penalizes the model for every elapsed time step, scaling higher when more occupants remain (Equation \ref{eq:r_time_penalty}). The hyperparameters $R_{\text{threat}}$, $R_{\text{exit}}$, $R_{\text{finished}}$, and $R_{\text{time}}$ weight these terms, providing fine-tuned control over the trade-off between rapid evacuation and threat exposure.

\subsection{The Graph Neural Network}
\label{sec:method_gnn}

Because raw features lack global context, node embeddings are essential for agents to evaluate a room's spatial relationship to threats, exits, and congestion points. We propose an edge-centric graph neural network that takes in raw node $\{v_i\}_{i\in V}$ and edge features $\{e_{i,j}\}_{(i,j)\in E}$ to generate node $\{h_i\}_{i\in V}$ and edge embeddings $\{w_{i,j}\}_{(i,j)\in E}$. 

We adopt an edge-first sequential update scheme (Equations \ref{edge_gnn} and \ref{node_gnn}). At each message-passing step, edge embeddings are updated by applying a two-layer MLP to the concatenation of the source node, previous edge, and target node embeddings. Subsequently, a GNN layer updates each node's embedding by aggregating information from its neighbors and the newly computed edges. While this framework supports various message-passing operators (Table \ref{tab:ablation_gpevac}), our primary model utilizes PNA. 

Each intermediate layer applies the ReLU activation function $\sigma$, augmented by residual connections and layer normalization to ensure stable training and better generalization. To support varying graph sizes, we pad the graph to a maximum node capacity, mask out non-existent nodes during the message-passing phase, and normalize all input features.

Let $h_i^{(k)}$ and $w_{i,j}^{(k)}$ denote the node and edge embeddings at layer $k$, initialized using the raw features as $h_i^{(0)}=v_i$ and $w_{i,j}^{(0)}=e_{i,j}$. We define the intermediate layer-normalized embeddings as $\tilde{h}_i^{(k)}=\text{LN}(h_i^{(k)})$ and $\tilde{w}_{i,j}^{(k)}=\text{LN}(w_{i,j}^{(k)})$. For each node $i$ with neighborhood $\mathcal{N}(i)$ and each edge $(i,j)$, the $K$-layer message passing sequence is formalized as:
\begin{align}
    w_{i,j}^{(k)}&=w_{i,j}^{(k-1)}+\sigma\left(\text{MLP}(\tilde{h}_i^{(k-1)}\parallel \tilde{w}_{i,j}^{(k-1)}\parallel \tilde{h}_j^{(k-1)})\right) \label{edge_gnn} \\
    h_i^{(k)} &= h_i^{(k-1)}+\sigma\left(\text{GNN}(\tilde{h}_i^{(k-1)}, \{\tilde{h}_j^{(k-1)}, w_{j,i}^{(k)}\}_{j\in \mathcal{N}(i)})\right)\label{node_gnn}
\end{align}
for $k=1,\dots,K$, where the final layer $K$ omits both the activation and the residual connections. We denote the output of the final layer $h_i^{(K)}$ and $w_{i,j}^{(K)}$ as $h_i$ and $w_{i,j}$ respectively.

Since graph representations of building layouts (Figure \ref{fig:example_graph})  have large diameters, long-range information is difficult to propagate through local messaging alone. To bypass this bottleneck while still maximizing expressiveness, we augment each graph with a virtual global node connected to all other nodes (Table \ref{tab:ablation_gpevac}). This collapses the distance between nodes to at most two hops, providing all nodes with information about distant threats and exits. To enable the network to differentiate between virtual and physical features, the initial global node $h_{\text{global}}^{(0)}$ and edge embeddings $\{w_{\text{global},i}^{(0)},w_{i,\text{global}}^{(0)}\}_{i\in V}$ are instantiated as learnable parameters.

\subsection{Proximal Policy Optimization}
\label{sec:ppo}

After generating the embeddings, GPEvac has sufficient information to estimate the optimal escape routes. Thus, at each timestep, the node embeddings are passed into a PPO algorithm with specialized critic $f_c$ and actor $f_a$ networks.

\subsubsection{Critic Network}
\label{sec:critic_network}

To encourage multi-agent cooperation for a globally optimal solution, we utilize a centralized critic:
\begin{align}
    f_c(G) = \text{MLP}(\overline{h} \parallel h_{\text{global}}) = \text{MLP}\left(\frac{1}{N}\sum_{i=1}^{N}h_i \Big\| h_{\text{global}}\right)\label{eq:critic_network}
\end{align}
To evaluate the holistic environment state rather than localized agent observations, the centralized critic aggregates graph-level features using a two-tier strategy. First, a global pooling layer computes the mean node embedding, $\bar{h}=\frac{1}{N}\sum_{i=1}^{N}h_i$, capturing the average local state. Second, to explicitly incorporate long-range structural dynamics, $\bar{h}$ is concatenated with the final global node embedding $h_{\text{global}}$. This unified embedding is passed through an MLP to compute the scalar state-value estimate $f_c(G)$.

\subsubsection{Actor Network}
\label{sec:actor_network}

During evacuation, each node acts as an independent agent for routing evacuees. GPEvac outputs a joint action vector $f_a(G)=[a_1,\dots,a_N]$, where each $a_i \in \mathcal{N}(i)$ denotes the routing decision for node $i$. Because standard multi-agent deep reinforcement learning suffers from instability and non-stationarity, we adopt a single-policy formulation that leverages the fully observable global state. 

A naïve centralized network with a fixed output dimension would introduce severe dependencies on arbitrary node orderings, failing to generalize across topologies. Instead, we implement an edge-centric scoring network that evaluates each individual node transition independently. This effectively quantifies the value of all possible actions across the graph, providing the logits for action selection.

Specifically, for any $(i,j)\in E$, the source node embedding $h_i$, the edge embedding $w_{i,j}$, and the target node embedding $h_j$ are concatenated and passed through an MLP to compute a local scalar routing score $s_{i,j}$ (Equation \ref{eq:actor_network}). These edge scores are then organized into an $N \times E_{\text{max}}$ matrix, where $E_{\text{max}}$ denotes the maximum number of outgoing edges for any node in the graph. In this matrix, rows correspond to source nodes and columns map to neighbor indices. 
\begin{align}
    s_{i,j}&=\text{MLP}(h_i\parallel w_{i,j}\parallel h_j), & (i,j)\in E\label{eq:actor_network} \\
    \pi_i(j)&=\frac{\exp(s_{i,j})}{\sum_{k\in \mathcal{N}(i)}\exp(s_{i,k})} & (i,j)\in E\label{eq:actor_softmax}
\end{align}

To obtain a valid probability distribution over available routing paths for each node, a masked softmax function is applied (Equation \ref{eq:actor_softmax}). This masking scheme operates on two distinct levels. First, to ensure the policy only evaluates decision-eligible nodes, it dynamically zeroes out rows corresponding to nodes that are empty or already routing evacuees. Second, to ensure that only valid actions are taken, within active rows it zeroes out columns that correspond to non-existent neighbor positions or blocked pathways.

\section{Results}
\label{sec:results}

We train GPEvac concurrently on both buildings (Figure \ref{fig:school-layouts}), utilizing 48 parallel simulation environments per layout. The PPO agent optimizes a shared policy across both topologies. To account for environmental stochasticity and properly assess generalization, we utilize a validation set and a test set. During training, we periodically evaluate the policy on the validation set to select the optimal model weights. The test set is used exclusively to assess final generalization. Both sets consist of 32 fixed-seed environments per layout, providing a robust sample size that ensures statistically reliable estimates of policy performance across diverse initial conditions.

To evaluate policy effectiveness in minimizing casualties, we compute the exposure time and cumulative threat penalty. Exposure time quantifies direct contact with the threat, calculated as the sum of the proportion of individuals occupying the shooter’s node at each timestep. The threat penalty (Equation \ref{eq_r_threat}) captures broader spatial risk by penalizing evacuees' proximity to the shooter, even if they are not at the exact same node. Alternative metrics, such as raw simulated casualty counts, depend on stochastic simulation mechanics and would introduce unnecessary evaluation noise.

As a secondary metric, we evaluate the overall evacuation time. Minimizing the duration of the event fundamentally reduces the window of vulnerability. This is directly quantified by the total episode length. Finally, we report the total episode return (Equation \ref{eq_r_t}) to provide a comprehensive scalar metric that captures the policy's ability to successfully balance rapid evacuation with spatial threat avoidance.

\subsection{Baseline Methods}
\label{sec:baselines}

Most existing evacuation models rely on predictable threat dynamics and struggle to scale to global multi-agent coordination. Furthermore, because no standardized benchmark environment exists for active-shooter scenarios, algorithms like ASTERS~\cite{LavalleRivera2023} and C-CASTERS~\cite{LavalleRivera2025} are tightly coupled to the specific simulations for which they were calibrated. Porting these models to our framework would require fundamentally altering their core mechanics and assumptions, rendering a direct head-to-head comparison scientifically invalid. Therefore, to ensure a rigorous and fair evaluation of our proposed architecture, we design and implement two highly competitive, intelligent baselines.

\subsubsection{Greedy Policy}
\label{sec:greedy_method}

The first baseline is a threat-oblivious greedy heuristic that directly minimizes total evacuation time. The rationale is that if individuals exit the environment as rapidly as possible, their cumulative vulnerability to the threat will also be low. To implement this, we utilize Dijkstra’s algorithm~\cite{dijkstra1959}, unconditionally directing evacuees to the next room along their shortest path to safety.

\subsubsection{Rule-Based Policy}
\label{sec:rule_based_method}

Our Rule-Based policy addresses the critical shortcoming of the greedy baseline by dynamically routing evacuees away from the threat. Inspired by the ``Run'' and ``Hide'' directives of the ``Run-Hide-Fight'' protocol, evacuees at node $i$ are routed according to Algorithm \ref{alg:rule-based-baseline}.

\begin{algorithm}[th]
\caption{Rule-Based Evacuation Policy}
\label{alg:rule-based-evacuation}
\textbf{Hyperparameter}: Distance threshold $\lambda$ \\
\textbf{Input}: Source node $i\in V$, eligible actions $\mathcal{A}=\mathcal{N}(i)\cup \{i\} \subseteq V$, set of exit nodes $V_{\text{exits}} \subseteq V$, and distance $d_v$ from each node $v\in V$ to the nearest shooter. \\
\textbf{Initialize}: Compute the shortest paths $P=\{p_1,p_2,p_3\}$ to the three nearest exits using Dijkstra's algorithm. \\
\textbf{Output}: Target node $j\in\mathcal{A}$
\begin{algorithmic}[1]
\STATE $d_{\text{safe}} \leftarrow \min(d_i, \lambda)$
\IF {$\mathcal{N}(i) \cap V_{\text{exits}} \neq \emptyset$}
    \STATE $j \leftarrow$ closest node in $\mathcal{N}(i) \cap V_{\text{exits}}$
\ELSIF {$\exists p_k \in P$ such that $d_v \geq d_{\text{safe}}$ for all $v \in p_k$}
    \STATE $j \leftarrow$ next node along the shortest qualifying $p_k$
\ELSE
    \STATE $j \leftarrow \arg\max_{v \in \mathcal{A}} d_v$
\ENDIF
\STATE \textbf{return} $j$
\end{algorithmic}
\label{alg:rule-based-baseline}
\end{algorithm}

At every time step, the policy chooses to escape or hide based on proximity to the shooter. The first two conditions of Algorithm \ref{alg:rule-based-baseline} define a ``Run'' strategy. If there exists a path to an exit where every node is farther from the shooter than the distance threshold, the policy selects the next node in that path. Otherwise, the policy ``hides'' by selecting whichever neighbor maximizes the distance to the nearest shooter.

This policy serves as a strong baseline because it yields near-optimal solutions and closely reflects human intuition. However, its performance depends on the choice of the safety threshold $\lambda$. To provide the strongest possible comparison, we empirically tune this baseline to its peak performance. Specifically, for each layout, we conduct a grid search over $\lambda$ using 100 simulation instances and select the threshold that maximizes the average cumulative reward.

\subsection{GPEvac}
\label{res:GPEvac}

Table \ref{tab:evaluation_metrics} 
compares GPEvac against the two baselines. Notably, a single policy was evaluated across both layouts, demonstrating the model's ability to capture complex crowd-threat dynamics across diverse topologies. A one-sided Wilcoxon signed-rank test ($N=64$) confirms that the best GPEvac policy significantly outperforms the Rule-Based baseline across all configurations, yielding lower exposure times ($p=3.65\times 10^{-2}$), reduced threat penalties ($p=6.71\times10^{-5}$), and higher returns ($p=4.92\times10^{-6}$).

\begin{table}[ht]
\centering
\small
\begin{tabular}{@{} l@{\hspace{4pt}} l @{\hspace{6pt}}|@{\hspace{6pt}} c@{\hspace{6pt}} c@{\hspace{6pt}} c@{\hspace{6pt}} c @{}}
    \toprule
    \textbf{Layout} & \textbf{Policy} & 
    \begin{tabular}{@{}c@{}}\textbf{Expos.} \\ \textbf{Time} ($\downarrow$)\end{tabular} &
    \begin{tabular}{@{}c@{}}\textbf{Threat} \\ \textbf{Pen.} ($\downarrow$)\end{tabular} & 
    \begin{tabular}{@{}c@{}}\textbf{Evac.} \\ \textbf{Time} ($\downarrow$)\end{tabular} & 
    \begin{tabular}{@{}c@{}}\textbf{Return} \\ ($\uparrow$)\end{tabular} \\
    \midrule

    \multirow{4}{*}{\begin{tabular}{@{}c@{}}\textbf{Acyclic} \\ \textbf{School} \end{tabular}} 
    & Greedy     & 1.383 & 12.56 & \textbf{83.75} & 6.15 \\
    & Rule-Based & 0.545 &  6.15 & 120.22 & 12.05 \\
    & GPEvac (Med.) & 0.579 & 6.19 & 107.13 & 12.10 \\
    & \textbf{GPEvac (Best)}   & \textbf{0.485} & \textbf{5.64} & 105.59 & \textbf{12.75} \\

    \midrule

    \multirow{4}{*}{\begin{tabular}{@{}c@{}}\textbf{Cyclic} \\ \textbf{School} \end{tabular}} 
    & Greedy     & 1.165 &  9.48 & \textbf{120.66} &  8.09 \\
    & Rule-Based & 0.491 &  4.23 & 145.97 & 12.94 \\
    & GPEvac (Med.) & 0.430 & 3.65 & 218.5 & 12.39 \\
    & \textbf{GPEvac (Best)}   & \textbf{0.370} & \textbf{3.08} & 158.5 & \textbf{13.94} \\

    \midrule

    \multirow{4}{*}{\begin{tabular}{@{}c@{}}\textbf{Average} \\ \textbf{(Both)} \end{tabular}} 
    & Greedy     & 1.274 &  11.02 & \textbf{102.20} &  7.12 \\
    & Rule-Based & 0.518 & 5.19 & 133.09 & 12.49 \\
    & GPEvac (Med.) & 0.500 & 4.91 & 177.39 & 12.29 \\
    & \textbf{GPEvac (Best)}   & \textbf{0.427} & \textbf{4.36} & 132.05 & \textbf{13.35} \\

    \bottomrule
\end{tabular}
\caption{Evaluation metrics for the proposed model and baselines, averaged across the test set of 32 environments per layout. The arrows ($\uparrow$ / $\downarrow$) indicate optimal direction. For GPEvac, we report the median performance across 10 independent training runs as well as the best-performing policy (selected as the one with the lowest threat penalty on the validation set). GPEvac achieves the lowest average exposure time and threat penalty, successfully minimizing contact with the active shooter.}
\label{tab:evaluation_metrics}
\end{table}

\begin{figure}[ht]
    \centering
    \includegraphics[width=0.71\columnwidth]{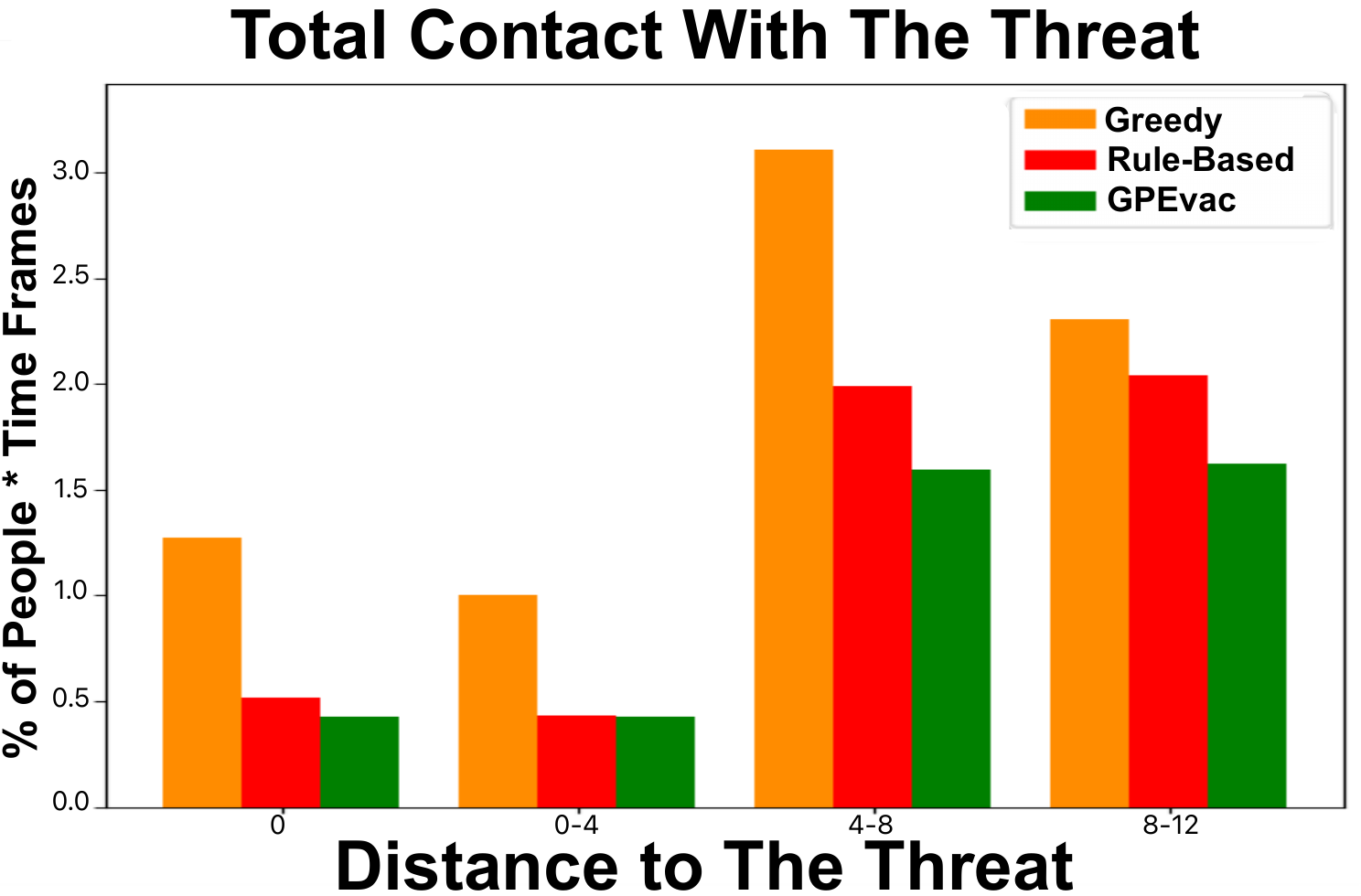}
    \caption{Total evacuee exposure categorized by distance to the active shooter. Bar heights represent the proportion of evacuees within each distance bin, aggregated over all time steps and averaged across 32 test environments per layout. A baseline level of immediate exposure (distance zero) is physically unavoidable due to the threat's randomized initialization.}
    \label{fig:distance_bars}
\end{figure}

To further quantify contact with the threat, we discretize the graph distances into bins and calculate the proportion of evacuees within each range at every time step. The cumulative exposure, averaged across all test environments, is presented in Figure \ref{fig:distance_bars}. GPEvac consistently achieves the lowest exposure across all proximity levels. This demonstrates that a learned policy can discover routing strategies that outperform intelligent human-designed heuristics, effectively justifying our deep reinforcement learning approach.

Qualitative observations of the policy rollouts (provided in the supplementary videos) reveal why the learned agent surpasses human-designed heuristics. GPEvac shares many foundational routing decisions with the Rule-Based approach, routing evacuees to exits on paths that maintain safe distances from the shooter. However, it further reduces contact with the threat by executing highly effective and sometimes non-intuitive maneuvers. For instance, rather than strictly hiding, GPEvac occasionally directs evacuees located 2-3 rooms away from the shooter to enter the hallway and begin evacuation. This proactive strategy prevents these evacuees from being cornered or trapped by the shooter's subsequent movements, minimizing overall threat exposure.

\subsection{Real-Time Inference}
\label{sec:real-time-inference}

Real-time operational capacity is a critical requirement for any evacuation system. Table \ref{tab:env_timing} evaluates the computational scalability of GPEvac. For each graph size, we generated 20 random graphs resembling typical building topologies and performed inference on randomly initialized weights. While environment initialization scales super-linearly due to distance matrix computations (Supplementary Material), it is strictly a one-time, offline preprocessing step. Crucially, single-step inference, which encompasses dynamic feature updates and the policy forward pass, scales sub-linearly.

\begin{table}[ht]
\centering
\begin{tabular}{lcccccc}
\toprule
\# Nodes & 50 & 100 & 200 & 400 & 800 & 1600 \\
\midrule
Init. (s) &0.03 &0.06 &0.16 &0.61 &3.13 &17.71 \\
Step (ms) &11.5 &16.0 &27.1 &43.1 &72.6 &126.2\\
\bottomrule
\end{tabular}
\caption{Environment creation times (s) and single-step inference times (ms) of GPEvac with varying-sized graphs on a single Apple M4 Pro CPU. Reported values are median times across 20 randomly initialized environments.}
\label{tab:env_timing}
\end{table}

In practice, latency for GPEvac is exceptionally low. For the Acyclic and Cyclic schools (55 and 70 nodes), GPEvac has a median inference time of just $14.73$ ms (95th percentile: $15.46$ ms), comfortably outpacing standard surveillance camera frame rates. Even for graphs exceeding 1,000 nodes, far larger than typical building layouts, inference executes in a fraction of a second. This means that GPEvac can deliver continuous, real-time routing updates in deployment without relying on high-end GPU hardware.



\subsection{Ablation Studies}
\label{res:ablation_studies}

\begin{table*}[t]
\centering
\small 
\setlength{\tabcolsep}{0pt} 
\begin{tabular*}{\textwidth}{@{\extracolsep{\fill}} l @{\hspace{0.2em}}| ccc @{\hspace{0.2em}}| ccc @{\hspace{0.2em}}| ccc @{}}
    \toprule
    \multirow{2}{*}{\textbf{Model Variant}} & 
    \multicolumn{3}{c}{\textbf{Acyclic School}} & 
    \multicolumn{3}{c}{\textbf{Cyclic School}} & 
    \multicolumn{3}{c}{\textbf{Average (Both)}} \\
    \cmidrule{2-4} \cmidrule{5-7} \cmidrule{8-10}
    
    & \begin{tabular}{@{}c@{}}\textbf{Threat} \\ \textbf{Pen.} ($\downarrow$)\end{tabular} 
    & \begin{tabular}{@{}c@{}}\textbf{Evacuation} \\ \textbf{Time} ($\downarrow$)\end{tabular} 
    & \begin{tabular}{@{}c@{}}\textbf{Return} \\ ($\uparrow$)\end{tabular} 
    
    & \begin{tabular}{@{}c@{}}\textbf{Threat} \\ \textbf{Pen.} ($\downarrow$)\end{tabular} 
    & \begin{tabular}{@{}c@{}}\textbf{Evacuation} \\ \textbf{Time} ($\downarrow$)\end{tabular}
    & \begin{tabular}{@{}c@{}}\textbf{Return} \\ ($\uparrow$)\end{tabular}
    
    & \begin{tabular}{@{}c@{}}\textbf{Threat} \\ \textbf{Pen.} ($\downarrow$)\end{tabular} 
    & \begin{tabular}{@{}c@{}}\textbf{Evacuation} \\ \textbf{Time} ($\downarrow$)\end{tabular}
    & \begin{tabular}{@{}c@{}}\textbf{Return} \\ ($\uparrow$)\end{tabular} \\
    \midrule
    
    \textbf{GPEvac} (with PNA) & 5.87 (0.29) & \textbf{100.1} (13.7) & \textbf{12.50} (0.24) & \textbf{2.97} (0.16) & \textbf{212.9} (98.3) & \textbf{13.09} (1.19) & \textbf{4.43} (0.21) & \textbf{172.2} (36.1) & \textbf{12.87} (0.89) \\
    \midrule
    
    \textbf{Structural Ablations} & & & & & & & & & \\
    \hspace{5pt} w/o Edge Updates   & 6.89 (4.53) & 122.9 (159.3) & 11.22 (4.89) & 3.77 (1.67) & 264.4 (141.4) & 11.14 (8.26) & 5.46 (2.93) & 227.5 (108.3) & 10.06 (8.19) \\
    \hspace{5pt} w/o Global Node    & 5.99 (0.55) & 102.1 (10.7) & 12.34 (0.54) & 3.26 (0.39) & 239.5 (116.0) & 12.82 (1.67) & 4.55 (0.25) & 172.2 (55.3) & 12.63 (1.04) \\
    \hspace{5pt} w/o Edge \& Global & 7.17 (1.37) & 143.1 (146.0) & 10.81 (3.31) & 3.84 (1.07) & 319.6 (158.9) & 9.32 (5.64) & 5.63 (1.22) & 234.9 (61.9) & 9.54 (4.33) \\
    \midrule
    
    \textbf{GNN Variants} & & & & & & & & & \\
    \hspace{5pt} ResGated          & \textbf{5.85} (0.81) & 113.1 (59.6) & 12.34 (1.11) & 3.42 (0.97) & 258.1 (172.7) & 10.56 (4.57) & 4.66 (1.09) & 208.9 (76.5) & 10.57 (2.99) \\
    \hspace{5pt} GATv2             & 5.90 (0.43) & 124.5 (32.3) & 12.35 (0.76) & 3.66 (0.61) & 251.0 (202.6) & 12.56 (4.52) & 4.81 (0.21) & 205.5 (104.1) & 12.04 (2.19) \\
    \hspace{5pt} Transformer       & 6.05 (0.78) & 112.5 (49.5) & 12.24 (1.13) & 3.80 (1.17) & 264.7 (97.4) & 11.75 (3.66) & 5.07 (1.36) & 188.8 (40.9) & 11.97 (2.51) \\
    \hspace{5pt} GINE              & 8.65 (16.9) & 282.4 (199.1) & 3.90 (35.5) & 5.17 (2.47) & 400.0 (30.0) & 4.13 (3.10) & 9.07 (8.30) & 341.2 (114.5) & 4.00 (16.97) \\
    \hspace{5pt} No GNN (MLP)             & 6.26 (0.60) & 142.6 (158.7) & 11.59 (4.12) & 4.00 (1.07) & 321.0 (51.9) & 7.31 (2.57) & 5.07 (0.67) & 231.4 (94.9) & 9.38 (3.69) \\
    \bottomrule
\end{tabular*}
\caption{Ablation studies of GPEvac across various model architectures and GNN layers. To ensure statistical robustness, each variant was trained independently 10 times. The model weights yielding the lowest threat penalty from each run were evaluated on the validation set of 32 environments per layout. Reported values represent the median across the 10 runs, with the interquartile range (IQR) in parentheses. Overall, GPEvac achieves the lowest contact with the threat when utilizing edge updates (Equation \ref{edge_gnn}), a virtual global node, and PNA as the GNN layer in Equation \ref{node_gnn}.}
\label{tab:ablation_gpevac}
\end{table*}

To isolate the contributions of individual components within GPEvac, we conduct a series of ablation studies (Table \ref{tab:ablation_gpevac}). First, to assess the impact of our message-passing scheme, we ablate the edge-update mechanism by substituting Equation \ref{edge_gnn} with the identity function. This significantly degrades the model's performance, elevating the threat penalty from 4.43 to 5.46 ($p=0.0186$) while inducing substantial variance across training runs. Moreover, eliminating the virtual global node from the message-passing layers and centralized critic (Equation \ref{eq:critic_network}) results in additional performance declines, highlighting its importance in global coordination.

To determine the most effective GNN layer for evacuation routing (Equation \ref{node_gnn}), we substitute PNA with alternative architectures. The GATv2, ResGated, and Graph Transformer variants achieve low threat penalties in the acyclic layout. However, they are much less consistent and their performance deteriorates in the more topologically complex cyclic layout. Additionally, the GINE was highly unstable, suggesting that not all aggregators work well for evacuation routing.

Removing the GNN entirely and replacing it with a two-layer MLP significantly increases contact with the threat and model variance. This confirms that explicitly encoding topological structure via message-passing is necessary for a globally optimal evacuation policy. We attribute PNA's superior performance to its use of multiple aggregators, which enables the network to simultaneously capture varying statistics such as average threat risk and minimum distance to the exit.

\section{Conclusion}
\label{sec:conclusion}

In this paper, we introduced GPEvac, a scalable, GNN-based reinforcement learning framework for dynamic evacuation routing during active shooting events. By representing architectural layouts as dynamic graphs and leveraging an edge-centric message-passing network with a virtual global node, this framework encodes critical long-range spatial relationships. Driven by a permutation-invariant edge-scoring actor, GPEvac overcomes the scalability limitations of prior work, allowing a single trained policy to outperform intelligent baselines across diverse building topologies.

The alarming escalation of mass shootings highlights the urgent need for dynamic public safety solutions. Capable of computing global routing updates in just 14.73 ms on local CPUs, GPEvac can readily integrate with existing surveillance networks equipped with real-time threat detection and tracking. By leveraging strategically placed digital screens and dynamic exit signs to actively steer occupants away from threats, this framework provides a practical pathway toward safer emergency evacuations.

Beyond real-time emergency routing, we envision GPEvac serving as a novel tool for urban planners. Simulating evacuation routing and comparing the exposure time across various layouts can provide a quantitative metric to evaluate the survivability of different building designs. Ultimately, GPEvac's unique architecture establishes a highly transferable foundation for dynamic routing in other complex domains, such as intelligent transportation and resource allocation.

\subsection{Limitations and Future Work}
\label{sec:limitations-future-work}

While GPEvac establishes a robust foundation, several avenues remain for future work. First, compiling a large-scale architectural dataset would enable rigorous training and evaluation for zero-shot generalization across diverse building topologies. Second, as detailed in the Supplementary Material, future iterations can incorporate more complex evacuation dynamics, such as capacity constraints, intelligent adversarial strategies, and stress-induced human behavior~\cite{Mawson1978PanicBehavior, Helbing2000SimulatingEscapePanic}.

Moreover, although modern tracking algorithms are highly accurate, chaotic real-world emergencies can introduce localization errors. Incorporating probabilistic node features would enable the evacuation policy to account for uncertainty in observational data. Finally, narrowing the simulation-to-reality gap for real-world deployment will require rigorous evaluation of GPEvac within high-fidelity simulation platforms such as Pedestrian Dynamics~\cite{incontrol_pedestrian_dynamics} or Unreal Engine~\cite{unreal_engine} to ensure operational safety and reliability.

\section*{Acknowledgments}

This work of Daniel Perkins was supported by the National Science Foundation Graduate Research Fellowship Program under Grant No. DGE-2146755.

\bibliography{aaai2027}

\begin{thebibliography}{56}
\providecommand{\natexlab}[1]{#1}

\bibitem[{Abouelaziz and Ghalmane(2026)}]{Abouelaziz2026DRAGON}
Abouelaziz, I.; and Ghalmane, Z. 2026.
\newblock DRAGON: A Dynamic Risk-Aware Graph Optimization Network for Adaptive Building Evacuation Using Graph Convolutional Network and Q-Learning.
\newblock \emph{Multimedia Tools and Applications}, 85(3): 197.

\bibitem[{Akbarzadeh et~al.(2025)Akbarzadeh, Moshashaei, Golzad, Liu, and Alizadeh}]{akbarzadeh2025visitor}
Akbarzadeh, O.; Moshashaei, P.; Golzad, H.; Liu, H.; and Alizadeh, S.~S. 2025.
\newblock Visitor responses to emergency evacuation: A human behavior approach.
\newblock \emph{Health Promotion Perspectives}, 15(4): 370--383.

\bibitem[{Almasan et~al.(2022)Almasan, Su{\'a}rez-Varela, Rusek, Barlet-Ros, and Cabellos-Aparicio}]{almasan2019deep}
Almasan, P.; Su{\'a}rez-Varela, J.; Rusek, K.; Barlet-Ros, P.; and Cabellos-Aparicio, A. 2022.
\newblock Deep Reinforcement Learning meets Graph Neural Networks: exploring a routing optimization use case.
\newblock In \emph{Computer Communications}.
\newblock Accepted version originally available as arXiv:1910.07421.

\bibitem[{Battaglia et~al.(2018)Battaglia, Hamrick, Bapst, Sanchez-Gonzalez, Zambaldi, Malinowski, Tacchetti, Raposo, Santoro, Faulkner, Gulcehre, Song, Ballard, Gilmer, Dahl, Vaswani, Allen, Nash, Langston, Dyer, Heess, Wierstra, Kohli, Botvinick, Vinyals, Li, and Pascanu}]{battaglia2018relational}
Battaglia, P.~W.; Hamrick, J.~B.; Bapst, V.; Sanchez-Gonzalez, A.; Zambaldi, V.; Malinowski, M.; Tacchetti, A.; Raposo, D.; Santoro, A.; Faulkner, R.; Gulcehre, C.; Song, F.; Ballard, A.; Gilmer, J.; Dahl, G.; Vaswani, A.; Allen, K.; Nash, C.; Langston, V.; Dyer, C.; Heess, N.; Wierstra, D.; Kohli, P.; Botvinick, M.; Vinyals, O.; Li, Y.; and Pascanu, R. 2018.
\newblock Relational inductive biases, deep learning, and graph networks.
\newblock \emph{arXiv preprint}.

\bibitem[{Bresson and Laurent(2017)}]{Bresson2017}
Bresson, X.; and Laurent, T. 2017.
\newblock Residual Gated Graph ConvNets.
\newblock \emph{arXiv}.

\bibitem[{Chen et~al.(2020)Chen, Lin, Li, Li, Zhou, and Sun}]{chen2020measuring}
Chen, D.; Lin, Y.; Li, W.; Li, P.; Zhou, J.; and Sun, X. 2020.
\newblock Measuring and Relieving the Over-smoothing Problem for Graph Neural Networks from the Topological View.
\newblock In \emph{Proceedings of the AAAI Conference on Artificial Intelligence}.

\bibitem[{Corso et~al.(2020)Corso, Cavalleri, Beaini, Li{\`o}, and Veli{\v{c}}kovi{\'c}}]{corso2020principal}
Corso, G.; Cavalleri, L.; Beaini, D.; Li{\`o}, P.; and Veli{\v{c}}kovi{\'c}, P. 2020.
\newblock Principal Neighbourhood Aggregation for Graph Nets.
\newblock In \emph{Advances in Neural Information Processing Systems}, volume~33, 13260--13271.

\bibitem[{Dijkstra(1959)}]{dijkstra1959}
Dijkstra, E.~W. 1959.
\newblock A note on two problems in connexion with graphs.
\newblock \emph{Numerische Mathematik}, 1(1): 269--271.

\bibitem[{{Epic Games}(2026)}]{unreal_engine}
{Epic Games}. 2026.
\newblock Unreal Engine.
\newblock Accessed: 2026-07-15.

\bibitem[{{FBI}(2014)}]{fbi_active_shooter_2014}
{FBI}. 2014.
\newblock A Study of Active Shooter Incidents in the United States Between 2000 and 2013.
\newblock Technical report, Federal Bureau of Investigation, U.S. Department of Justice.

\bibitem[{Freeman(1978)}]{freeman1978centrality}
Freeman, L.~C. 1978.
\newblock Centrality in Social Networks: Conceptual Clarification.
\newblock \emph{Social Networks}, 1(3): 215--239.

\bibitem[{Gilmer et~al.(2017)Gilmer, Schoenholz, Riley, Vinyals, and Dahl}]{gilmer2017neural}
Gilmer, J.; Schoenholz, S.~S.; Riley, P.~F.; Vinyals, O.; and Dahl, G.~E. 2017.
\newblock Neural Message Passing for Quantum Chemistry.
\newblock In \emph{Proceedings of the 34th International Conference on Machine Learning}, volume~70, 1469--1478. PMLR.

\bibitem[{Gunn et~al.(2017)Gunn, Luh, Lu, and Hotaling}]{Gunn2017}
Gunn, S.; Luh, P.~B.; Lu, X.; and Hotaling, B. 2017.
\newblock Optimizing guidance for an active shooter event.
\newblock In \emph{2017 IEEE International Conference on Robotics and Automation (ICRA)}, 4299--4304.

\bibitem[{Hart, Nilsson, and Raphael(1968)}]{hart1968formal}
Hart, P.~E.; Nilsson, N.~J.; and Raphael, B. 1968.
\newblock A Formal Basis for the Heuristic Determination of Minimum Cost Paths.
\newblock \emph{IEEE Transactions on Systems Science and Cybernetics}, 4(2): 100--107.

\bibitem[{Helbing, Farkas, and Vicsek(2000)}]{Helbing2000SimulatingEscapePanic}
Helbing, D.; Farkas, I.~J.; and Vicsek, T. 2000.
\newblock Simulating dynamical features of escape panic.
\newblock \emph{Nature}, 407(6803): 487--490.

\bibitem[{Hu et~al.(2021)Hu, Fey, Zitnik, Dong, Ren, Liu, Catasta, and Leskovec}]{hu2020open}
Hu, W.; Fey, M.; Zitnik, M.; Dong, Y.; Ren, H.; Liu, B.; Catasta, M.; and Leskovec, J. 2021.
\newblock Open Graph Benchmark: Datasets for Machine Learning on Graphs.
\newblock \emph{Proceedings of the National Academy of Sciences}, 118(12): e2111777118.

\bibitem[{Huang et~al.(2023)Huang, Liang, Xiao, Fang, Li, and Ye}]{Huang2023DynamicScanning}
Huang, Z.; Liang, R.; Xiao, Y.; Fang, Z.; Li, X.; and Ye, R. 2023.
\newblock Simulation of pedestrian evacuation with reinforcement learning based on a dynamic scanning algorithm.
\newblock \emph{Physica A: Statistical Mechanics and its Applications}, 625: 129011.

\bibitem[{{InControl}(2026)}]{incontrol_pedestrian_dynamics}
{InControl}. 2026.
\newblock Pedestrian Dynamics.
\newblock Accessed: 2026-07-15.

\bibitem[{Jaderberg et~al.(2018)Jaderberg, Czarnecki, Dunning, Marris, Lever, Casta{\~n}eda, Beattie, Rabinowitz, Morcos, Ruderman, Sonnerat, Green, Deason, Leibo, Silver, Hassabis, Kavukcuoglu, and Graepel}]{Jaderberg2019}
Jaderberg, M.; Czarnecki, W.~M.; Dunning, I.; Marris, L.; Lever, G.; Casta{\~n}eda, A.~G.; Beattie, C.; Rabinowitz, N.~C.; Morcos, A.~S.; Ruderman, A.; Sonnerat, N.; Green, T.; Deason, L.; Leibo, J.~Z.; Silver, D.; Hassabis, D.; Kavukcuoglu, K.; and Graepel, T. 2018.
\newblock Human-level performance in 3D multiplayer games with population-based reinforcement learning.
\newblock \emph{Science}, 364: 859--865.

\bibitem[{Kipf and Welling(2017)}]{Kipf2017}
Kipf, T.~N.; and Welling, M. 2017.
\newblock Semi-Supervised Classification with Graph Convolutional Networks.
\newblock In \emph{5th International Conference on Learning Representations, {ICLR} 2017}. OpenReview.net.

\bibitem[{Klein and Bryant(2025)}]{klein2025gunviolencearchive}
Klein, M.; and Bryant, M. 2025.
\newblock Gun Violence Archive.
\newblock \url{https://www.gunviolencearchive.org}.
\newblock Accessed: 2026-04-26.

\bibitem[{Kotecha and del Rio~Chanona(2025)}]{Kotecha2025}
Kotecha, N.; and del Rio~Chanona, A. 2025.
\newblock Leveraging graph neural networks and multi-agent reinforcement learning for inventory control in supply chains.
\newblock \emph{Computers \& Chemical Engineering}, 186: 109111.

\bibitem[{Lavalle-Rivera et~al.(2023)Lavalle-Rivera, Ramesh, Harris, and Chakraborty}]{LavalleRivera2023}
Lavalle-Rivera, J.; Ramesh, A.; Harris, L.~M.; and Chakraborty, S. 2023.
\newblock The effectiveness of naive optimization of the egress path for an active-shooter scenario.
\newblock \emph{Heliyon}, 9(2): e13695.

\bibitem[{Lavalle‐Rivera, Ramesh, and Chakraborty(2026)}]{LavalleRivera2025}
Lavalle‐Rivera, J.; Ramesh, A.; and Chakraborty, S. 2026.
\newblock An Optimized Evacuation Plan for an Active‐Shooter Situation Constrained by Network Capacity.
\newblock \emph{Journal of Contingencies and Crisis Management}, 34.

\bibitem[{Lee(2019)}]{Lee2019RunHideFightABM}
Lee, J.~Y. 2019.
\newblock Agent-Based Modeling to Assess the Effectiveness of Run Hide Fight.

\bibitem[{Li et~al.(2024)Li, Zhang, Alizadeh, Zhang, Duffield, Meyer et~al.}]{li2023reinforcement}
Li, D.; Zhang, Z.; Alizadeh, B.; Zhang, Z.; Duffield, N.; Meyer, M.~A.; et~al. 2024.
\newblock A reinforcement learning-based routing algorithm for large street networks.
\newblock \emph{International Journal of Geographical Information Science}, 38(2): 183--215.

\bibitem[{Liu and Meidani(2025)}]{Liu2025}
Liu, T.; and Meidani, H. 2025.
\newblock Graph Neural Networks for Travel Distance Estimation and Route Recommendation Under Probabilistic Hazards.
\newblock \emph{International Journal of Transportation Science and Technology}, 21.

\bibitem[{Maggiolino et~al.(2023)Maggiolino, Ahmad, Cao, and Kitani}]{maggiolino2023deepocs}
Maggiolino, G.; Ahmad, A.; Cao, J.; and Kitani, K. 2023.
\newblock Deep {OC-SORT}: Multi-Pedestrian Tracking by Adaptive Re-Identification.
\newblock \emph{arXiv preprint arXiv:2302.11813}.

\bibitem[{Mawson(1978)}]{Mawson1978PanicBehavior}
Mawson, A. 1978.
\newblock Panic behavior: a review and a new hypothesis.
\newblock In \emph{Proceedings of the 9th World Congress of Sociology}. Uppsala, Sweden.
\newblock Chalfont Research Institute.

\bibitem[{Medina and Drake(2026)}]{Medina2026GNNPolymers}
Medina, H.; and Drake, R. 2026.
\newblock Graph Neural Networks for Polymer Characterization and Property Prediction: Opportunities and Challenges.
\newblock \emph{Journal of Chemical Information and Modeling}, 66(3): 1316--1336.

\bibitem[{Mnih et~al.(2016)Mnih, Badia, Mirza, Graves, Harley, Lillicrap, Silver, and Kavukcuoglu}]{Mnih2016}
Mnih, V.; Badia, A.~P.; Mirza, M.; Graves, A.; Harley, T.; Lillicrap, T.~P.; Silver, D.; and Kavukcuoglu, K. 2016.
\newblock Asynchronous methods for deep reinforcement learning.
\newblock In \emph{Proceedings of the 33rd International Conference on International Conference on Machine Learning - Volume 48}, ICML'16, 1928–1937. JMLR.org.

\bibitem[{Mnih et~al.(2013)Mnih, Kavukcuoglu, Silver, Graves, Antonoglou, Wierstra, and Riedmiller}]{Mnih2013}
Mnih, V.; Kavukcuoglu, K.; Silver, D.; Graves, A.; Antonoglou, I.; Wierstra, D.; and Riedmiller, M. 2013.
\newblock Playing {Atari} with {Deep} Reinforcement {Learning}.
\newblock In \emph{Advances in {Neural} Information Processing Systems 26 (NIPS 2013 Deep Learning Workshop)}. Lake Tahoe, NV, USA.
\newblock ArXiv:1312.5602.

\bibitem[{{National Center for Education Statistics}(2026)}]{nces_ntps_2020_21_classsize}
{National Center for Education Statistics}. 2026.
\newblock Average public school class size: Average class size in public K-12 schools, by school level, class type, and state: 2020-21.
\newblock \url{https://nces.ed.gov/surveys/ntps/estable/table/ntps/ntps2021_sflt07_t1s}.
\newblock U.S. Department of Education, National Teacher and Principal Survey (NTPS), ``Public School Teacher Data File,'' 2020--21.

\bibitem[{Papoudakis et~al.(2019)Papoudakis, Christianos, Rahman, and Albrecht}]{Papoudakis2019DealingWN}
Papoudakis, G.; Christianos, F.; Rahman, A.; and Albrecht, S.~V. 2019.
\newblock Dealing with Non-Stationarity in Multi-Agent Deep Reinforcement Learning.
\newblock \emph{ArXiv}, abs/1906.04737.

\bibitem[{Redmon et~al.(2016)Redmon, Divvala, Girshick, and Farhadi}]{redmon2016yolo}
Redmon, J.; Divvala, S.; Girshick, R.; and Farhadi, A. 2016.
\newblock You Only Look Once: Unified, Real-Time Object Detection.
\newblock \emph{arXiv preprint arXiv:1506.02640}.

\bibitem[{Robinson et~al.(2026)Robinson, Robicheaux, Popov, Ramanan, and Peri}]{robinson2026rfdetr}
Robinson, I.; Robicheaux, P.; Popov, M.; Ramanan, D.; and Peri, N. 2026.
\newblock {RF-DETR}: Neural Architecture Search for Real-Time Detection Transformers.
\newblock In \emph{International Conference on Learning Representations (ICLR)}.
\newblock Accepted at ICLR 2026.

\bibitem[{Rosa et~al.(2023)Rosa, Falqueiro, Bonacin, Mendonça, Rodrigo~Filho, and Gonçalves}]{Rosa2023EvacuAI}
Rosa, A.~C.; Falqueiro, M.~C.; Bonacin, R.; Mendonça, F. L. L.~d.; Rodrigo~Filho, G.~P.; and Gonçalves, V.~P. 2023.
\newblock EvacuAI: An Analysis of Escape Routes in Indoor Environments with the Aid of Reinforcement Learning.
\newblock \emph{Sensors}, 23(21): 8892.

\bibitem[{Rummaneethorn, Taubman, and Adamakos(2024)}]{Rummaneethorn2024RunHideFightHospital}
Rummaneethorn, N.; Taubman, C.; and Adamakos, F. 2024.
\newblock Is the Run, Hide, Fight Concept Effective in Improving Hospital Response to Shooting Incidents? A Systematic Review.
\newblock \emph{Disaster Medicine and Public Health Preparedness}, 18: e111.

\bibitem[{Schulman et~al.(2017)Schulman, Wolski, Dhariwal, Radford, and Klimov}]{Schulman2017}
Schulman, J.; Wolski, F.; Dhariwal, P.; Radford, A.; and Klimov, O. 2017.
\newblock Proximal Policy Optimization Algorithms.
\newblock \emph{CoRR}, abs/1707.06347.

\bibitem[{Shao et~al.(2021)Shao, Li, Wu, Zhifeng, and Zhang}]{Shao2021}
Shao, Y.; Li, R.; Wu, Y.; Zhifeng, Z.; and Zhang, H. 2021.
\newblock Graph Attention Network-Based Multi-Agent Reinforcement Learning for Slicing Resource Management in Dense Cellular Network.
\newblock \emph{IEEE Transactions on Vehicular Technology}, PP: 1--1.

\bibitem[{Sharma et~al.(2021)Sharma, Fernandez, Zaroukian, Dorothy, Basak, and Asher}]{Sharma2021SurveyMARL}
Sharma, P.~K.; Fernandez, R.; Zaroukian, E.; Dorothy, M.; Basak, A.; and Asher, D.~E. 2021.
\newblock Survey of recent multi-agent reinforcement learning algorithms utilizing centralized training.
\newblock In \emph{Artificial Intelligence and Machine Learning for Multi-Domain Operations Applications III}, volume 11746, 117462K. SPIE.

\bibitem[{Tokhi(2017)}]{tokhi2017runhidefight}
Tokhi, M. 2017.
\newblock Run, Hide, Fight Comes to Dominion.

\bibitem[{{U.S. Department of Homeland Security}(2026)}]{DHSActiveShooterPreparedness}
{U.S. Department of Homeland Security}. 2026.
\newblock {Active Shooter Preparedness Resources}.
\newblock \url{https://www.dhs.gov/active-shooter-preparedness}.
\newblock Includes ``Run. Hide. Fight.'' training materials.

\bibitem[{Velickovic et~al.(2018)Velickovic, Cucurull, Casanova, Romero, Li{\`o}, and Bengio}]{Velickovic2018}
Velickovic, P.; Cucurull, G.; Casanova, A.; Romero, A.; Li{\`o}, P.; and Bengio, Y. 2018.
\newblock Graph Attention Networks.
\newblock In \emph{International Conference on Learning Representations (ICLR)}, volume abs/1710.10903.

\bibitem[{Waite et~al.(2023)Waite, Feng, Tavassoli, Harris, Tan, Chakraborty, and Sarkar}]{Waite2023ActiveSD}
Waite, J.~R.; Feng, J.; Tavassoli, R.; Harris, L.~M.; Tan, S.~Y.; Chakraborty, S.; and Sarkar, S. 2023.
\newblock Active shooter detection and robust tracking utilizing supplemental synthetic data.
\newblock \emph{ArXiv}, abs/2309.03381.

\bibitem[{Weisstein(2026)}]{mathworld_graph_eccentricity}
Weisstein, E.~W. 2026.
\newblock Graph Eccentricity.
\newblock \url{https://mathworld.wolfram.com/GraphEccentricity.html}.
\newblock MathWorld--A Wolfram Web Resource.

\bibitem[{Wu and Mu(2025)}]{wu2025toward}
Wu, S.; and Mu, R. 2025.
\newblock Toward Sustainable and Inclusive Cities: Graph Neural Network-Enhanced Optimization for Disability-Inclusive Emergency Evacuation in High-Rise Buildings.
\newblock \emph{Sustainability}, 17(22): 10387.

\bibitem[{Xu et~al.(2020)Xu, Huang, Mango, Li, and Li}]{xu2020simulating}
Xu, D.; Huang, X.; Mango, J.; Li, X.; and Li, Z. 2020.
\newblock Simulating multi-exit evacuation using deep reinforcement learning.
\newblock \emph{Transactions in GIS}.
\newblock Preprint submitted to Transactions in GIS.

\bibitem[{Xu et~al.(2019)Xu, Hu, Leskovec, and Jegelka}]{xu2018how}
Xu, K.; Hu, W.; Leskovec, J.; and Jegelka, S. 2019.
\newblock How Powerful are Graph Neural Networks?
\newblock In \emph{International Conference on Learning Representations}.

\bibitem[{Yu et~al.(2022)Yu, Velu, Vinitsky, Gao, Wang, Bayen, and Wu}]{Yu2022}
Yu, C.; Velu, A.; Vinitsky, E.; Gao, J.; Wang, Y.; Bayen, A.; and Wu, Y. 2022.
\newblock The surprising effectiveness of PPO in cooperative multi-agent games.
\newblock In \emph{Proceedings of the 36th International Conference on Neural Information Processing Systems}, NIPS '22. Red Hook, NY, USA: Curran Associates Inc.
\newblock ISBN 9781713871088.

\bibitem[{Yun et~al.(2019)Yun, Jeong, Kim, Kang, and Kim}]{yun2019graph}
Yun, S.; Jeong, M.; Kim, R.; Kang, J.; and Kim, H.~J. 2019.
\newblock Graph Transformer Networks.
\newblock In \emph{Advances in Neural Information Processing Systems (NeurIPS)}.

\bibitem[{Zeng(2021)}]{Zeng2022}
Zeng, Z. 2021.
\newblock GraphLight: Graph-based Reinforcement Learning for Traffic Signal Control.
\newblock In \emph{2021 IEEE 6th International Conference on Computer and Communication Systems (ICCCS)}, 645--650.

\bibitem[{Zhang, Chai, and Lykotrafitis(2021)}]{Zhang2021DeepReinforcement}
Zhang, Y.; Chai, Z.; and Lykotrafitis, G. 2021.
\newblock Deep reinforcement learning with a particle dynamics environment applied to emergency evacuation of a room with obstacles.
\newblock \emph{Physica A: Statistical Mechanics and its Applications}, 571: 125845.

\bibitem[{Zhang, Yang, and Zhu(2024)}]{zhang2024double}
Zhang, Y.; Yang, B.; and Zhu, J. 2024.
\newblock A double-layer crowd evacuation simulation method based on deep reinforcement learning.
\newblock \emph{Computer Animation and Virtual Worlds}, 35(3): e2280.

\bibitem[{Zhang, Tan, and Tiong(2025)}]{Zhang2025}
Zhang, Z.; Tan, L.; and Tiong, R. 2025.
\newblock Evacuation path optimization algorithm for grassland fires based on SAR imagery and intelligent optimization.
\newblock \emph{Frontiers in Environmental Science}, 13.

\bibitem[{Zhao et~al.(2024)Zhao, Huo, Li, Feng, Yu, Qi, and Wang}]{Zhao2025}
Zhao, B.; Huo, M.; Li, Z.; Feng, W.; Yu, Z.; Qi, N.; and Wang, S. 2024.
\newblock Graph-based multi-agent reinforcement learning for collaborative search and tracking of multiple UAVs.
\newblock \emph{Chinese Journal of Aeronautics}, 38.

\end{thebibliography}
\setcounter{secnumdepth}{2} 
\clearpage
\twocolumn[
\begin{center}
  {\LARGE\bfseries Appendices}
\end{center}
\vspace{0.5em}
]
\appendix

\section{Introduction and Related Works}

\begin{figure}[h]
    \centering
    \includegraphics[width=0.8\columnwidth]{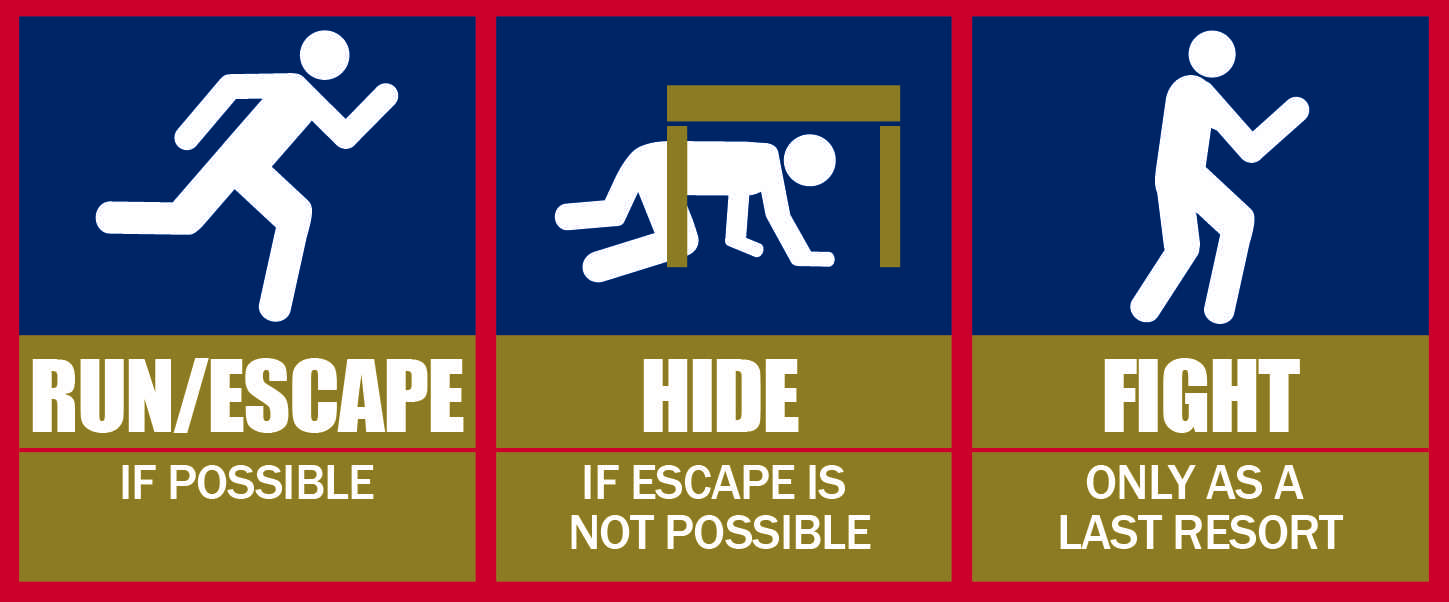}
    \caption{The ``Run-Hide-Fight'' Protocol \cite{tokhi2017runhidefight}}
    \label{fig:rhf}
\end{figure}

As discussed in the introduction, the Department of Homeland Security's ``Run-Hide-Fight'' protocol serves as the standard framework for active-shooter response. Under this protocol, individuals make decentralized decisions based on their immediate surroundings, ultimately choosing whether to evacuate, stay in place, or confront the attacker. This decision-making process is illustrated in Figure \ref{fig:rhf}.

\section{Methodology}
\label{app:methodology}


\subsection{Dataset}
\label{app:dataset}
To train and validate GPEvac, we utilize two school building layouts adapted from the C-CASTERS framework \cite{LavalleRivera2025}. These architectural layouts are abstracted into graph representations, where nodes represent individual rooms and edges denote connecting passageways. To capture realistic movement dynamics, we use the edge weights computed in this prior work, which represent traversal times empirically derived via Unreal Engine~\cite{unreal_engine} simulations. 

\subsection{Node and Edge Features}
\label{app:node_edge_features}
To provide GPEvac with the necessary context to learn optimal routing policies, we construct several hand-crafted features based on the layout topology, the shooter's location, and the distribution of evacuees. To ensure model robustness, these features are deliberately redundant. We categorize the 47 node features into two main groups:

\begin{itemize}
    \item \textbf{Static Features:} These features are computed once during initialization and stay constant throughout the entire simulation.
    \begin{itemize}
        \item \textbf{\textit{exit\_mask} and \textit{hallway\_mask}:} Boolean variables that indicate whether or not a node is an exit or a hallway.
        \item \textbf{\textit{num\_neighbors\_within\_[x]\_hops}:} Four different features indicating the number of nodes within $x\in{\{1,2,3,4\}}$ hops.
        \item  \textbf{\textit{num\_exit\_nodes\_within\_[x]\_hops}:} Four features indicating the number of exit nodes within $x$ hops.
        \item \textbf{\textit{proportion\_neigbors\_hallway}:} The percentage of neighbors that are hallways.
        \item  \textbf{\textit{distance\_to\_exit\_q[x]}:} Three boolean features indicating whether or not the current node is within a certain quartile distance to the nearest exit.
        \item \textbf{\textit{closeness\_centrality}:} How near a node is to all other nodes, via shortest-path distances~\cite{freeman1978centrality}.
        \item \textbf{\textit{betweenness\_centrality}:}  How often the node lies on shortest paths between other nodes in the graph~\cite{freeman1978centrality}.
        \item \textbf{\textit{eccentricty}:} Maximum graph distance between the current node and any other node~\cite{mathworld_graph_eccentricity}
        \item \textbf{\textit{min\_edge\_weight} and \textit{avg\_edge\_weight}:} The minimum and average edge weights (travel time) of all connected edges.
        \item  \textbf{\textit{distance\_to\_[x]\_exit}:} Three real number features indicating the distance to the $x$'th nearest exit.
        \item  \textbf{\textit{num\_hops\_to\_[x]\_exit}:} Three integer features indicating the number of hops to the $x$'th nearest exit.
    \end{itemize}
    
    \item \textbf{Dynamic Features:} These features are updated at each timestep, as they depend on the location of the shooters and evacuees.
    \begin{itemize}
        \item \textbf{\textit{num\_people}:} The current number of people in the node.
        \item \textbf{\textit{congestion}:} A weighted sum of the number of people within 1, 2, and 3 hops (using the \textit{congestion\_weights} hyperparameter).
        \item  \textbf{\textit{num\_people\_in\_[x]\_hop\_neighbors}:} The number of people exactly $x\in\{1,2,3\}$ hops away.
        \item  \textbf{\textit{num\_people\_within\_[x]\_hop\_neighbors}:} The number of people within $x\in\{1,2,3\}$ hops.
        \item  \textbf{\textit{max\_num\_people\_[x]\_hops}:} The maximum number of people for all the nodes $x\in\{1,2\}$ hops away.
        \item \textbf{\textit{num\_threats}:} The number of threats in the current node.
        \item  \textbf{\textit{distance\_to\_[x]\_threat}:} The distance to the $x\in\{1,2,3\}$ nearest threat.
        \item  \textbf{\textit{num\_hops\_to\_[x]\_threat}:} The number of hops to the $x\in\{1,2,3\}$ nearest threat.
        \item  \textbf{\textit{num\_threats\_in\_[x]\_hop}:} The number of threats exactly $x\in\{1,2,3\}$ hops away.
        \item  \textbf{\textit{num\_threats\_within\_[x]\_hop}:} The number of threats within $x\in\{1,2,3\}$ hops.
    \end{itemize}
\end{itemize}

Similarly, GPEvac uses the following 20 edge features:

\begin{itemize}
    \item \textbf{Static Features:} These features are computed once during initialization and stay constant throughout the entire simulation.
    \begin{itemize}
        \item \textbf{\textit{weight}:} The original edge weight defined from C-CASTERS~\cite{LavalleRivera2025}. This indicates the estimated travel time between nodes.
        \item \textbf{\textit{in\_path\_to\_nearest\_exit}:} A boolean variable indicating whether or not this edge is along the shortest path from the source node to the nearest exit.
        \item  \textbf{\textit{delta\_distance\_to\_[x]\_exit}:} The difference in distances between the source node and target node to each of their $x\in\{1,2\}$ nearest exits.
        \item  \textbf{\textit{delta\_num\_hops\_to\_[x]\_exit}:} The difference in number of hops between the source node and target node to each of their $x\in\{1,2\}$ nearest exits.
    \end{itemize}
    
    \item \textbf{Dynamic Features:} These features are updated at each timestep, as they depend on the location of the shooters and evacuees.
    \begin{itemize}
        \item \textbf{\textit{num\_people}:} The current number of people in transit from the source node to the target node. 
        \item \textbf{\textit{time\_steps\_left}:} The number of timesteps left before the current action is complete (the hyperparameter \textit{max\_num\_steps\_per\_action} controls how many steps each action takes).
        \item \textbf{\textit{delta\_num\_people}:} The difference in the number of people between the source node and target node.
        \item \textbf{\textit{delta\_congestion}:} The difference in congestion between the source node and target node.
        \item  \textbf{\textit{delta\_num\_people\_within\_[x]\_hops}:} The difference in \textit{num\_people\_in\_[x]\_hop\_neighbors} between the source node and target node for $x\in\{1,2\}$.
        \item \textbf{\textit{delta\_num\_threats}:} The difference in the number of threats between the source node and target node.
        \item  \textbf{\textit{delta\_distance\_to\_[x]\_threat}:} The difference in \textit{distance\_to\_[x]\_threat} between the source node and target node for $x\in\{1,2\}$.
        \item  \textbf{\textit{delta\_num\_hops\_to\_[x]\_threat}:} The difference in \textit{num\_hops\_to\_[x]\_threat} between the source node and target node for $x\in\{1,2\}$.
        \item  \textbf{\textit{delta\_num\_threats\_within\_[x]\_hops}:} The difference in \textit{num\_threats\_within\_[x]\_hop} between the source node and target node for $x\in\{1,2,3\}$.
    \end{itemize}
\end{itemize}

\subsection{Temporal Complexity}
\label{app:temporal-complexity}

As demonstrated in Table \ref{tab:env_timing}, GPEvac's initialization time scales super-linearly with the number of nodes, whereas single-step inference scales sub-linearly. This disparity is an intentional architectural design: because rapid decision-making is critical during an active shooting, we shift the computational overhead from real-time execution to an offline initialization phase. During deployment, latency is primarily driven by only two operations: (1) feature extraction and (2) the policy forward pass.

To minimize real-time computational overhead when extracting node and edge features, we preprocess the building graphs to compute the shortest paths between all pairs of nodes. By running Dijkstra's algorithm \cite{dijkstra1959} from every node, we construct a complete distance routing matrix. Because public spaces in our formulation are unlikely to contain more than a thousand nodes, this preprocessing step remains computationally feasible for practically any building layout. Computing the all-pairs shortest paths operates in $O(\vert{}V\vert{} (\vert{}V\vert{} + \vert{}E\vert{}) \log \vert{}V\vert{})$ time, and caching the resulting distance matrix requires only $O(\vert{}V\vert{}^2)$ memory space, where $\vert{}V\vert{}$ and $\vert{}E\vert{}$ denote the number of nodes and edges. Ultimately, this makes real-time feature extraction possible, allowing GPEvac to update distance-based features for all nodes in $O(\vert{}V\vert{})$ using only $O(1)$ lookups.

During the policy forward pass, the message-passing architecture requires each node to aggregate features from its local neighborhood. Assuming a sparse building topology where the average number of edges per node is bounded, this operation scales linearly with the number of nodes $O(\vert{}V\vert{})$. Similarly, the centralized critic pools all node embeddings to estimate the global state, requiring an additional $O(\vert{}V\vert{})$ operations. Finally, the actor network evaluates independent routing scores for each available pathway, contributing a computational cost proportional to the total number of edges $O(\vert{}E\vert{})$. Combining these components, while keeping the dimension of the hidden layers constant, yields an asymptotic complexity of $O(\vert{}V\vert{} + \vert{}E\vert{})$. As observed in Section \ref{sec:real-time-inference}, the empirical latency scales even more efficiently, exhibiting sub-linear growth.

\subsubsection{Synthetic Graph Generation}
\label{app:synthetic_graph}
To rigorously benchmark the inference times presented in Table \ref{tab:env_timing}, we evaluated GPEvac across 20 synthetic graphs for each target node count ($N$). Rather than relying on random edge assignments, these graphs were procedurally generated to mimic the topological constraints of physical building floor plans. The number of exit nodes scales logarithmically with the total node count, perturbed by a stochastic variance of $\pm 1$. Of the remaining nodes, approximately $35\%$ are designated as hallways. To ensure realistic pathways, hallway nodes are linked into a continuous spine with a small number of cross-links. Exits are evenly distributed along this spine, and individual rooms are attached as degree-1 stubs to randomly selected hallways. To enforce physical spatial limitations, node degrees are strictly capped at $6$. This procedure ensures the resulting topologies maintain realistic degree distributions and edge-to-node ratios while being parameterized by $N$.

\subsection{Simulation Environment}
\label{app:sim}

As detailed in Section \ref{sec:sim_env}, the simulation models evacuation as a discretized flow across a building graph. To enhance realism, evacuees traverse edges incrementally rather than instantly teleporting between nodes. The traversal capacity, the maximum number of people who can cross an edge within a fixed time horizon $T$ 
by the edge weight. Consequently, higher edge weights yield lower flow capacities. This total capacity is then distributed evenly across the $T$ timesteps. If the number of waiting evacuees is lower than the edge's capacity, the simulation schedules only those present, meaning their crossing can conclude in fewer than $T$ steps. Finally, to best represent human reaction time, once evacuees enter transit, they are committed to their destination; the simulation prevents mid-edge reassignment until the allocation is complete.

Although GPEvac lacks an explicit penalty function for congestion, this discretized movement organically generates an implicit learning signal. Because edge weights dictate the maximum throughput of evacuees exiting a node over $T$ timesteps, structural bottlenecks naturally penalize the model. When a node's population exceeds its egress capacity, the resulting congestion throttles evacuation flow. This delay inherently increases the accumulated time-step penalties and prolongs evacuee exposure to active threats, naturally guiding the policy to learn bottleneck avoidance.

\subsection{Normalization}
\label{app:normalization}

To ensure stable and efficient training across diverse graph sizes and topologies, GPEvac applies feature-specific normalization. For example, graph topology metrics, such as node degree, are normalized by the maximum allowed degree, while distance-based features are normalized by the graph's maximum distance or hops to an exit. To avoid outliers, many of these features are subsequently clipped to an upper bound of 2. Furthermore, to provide useful information to both the actor and critic networks, localized evacuee counts are normalized in two different ways: once by a predefined node capacity (50) and again by the total number of people in the graph.

\subsection{Virtual Global Node}
\label{app:global_node}

\begin{figure}[h]
    \centering
    \includegraphics[width=1\columnwidth]{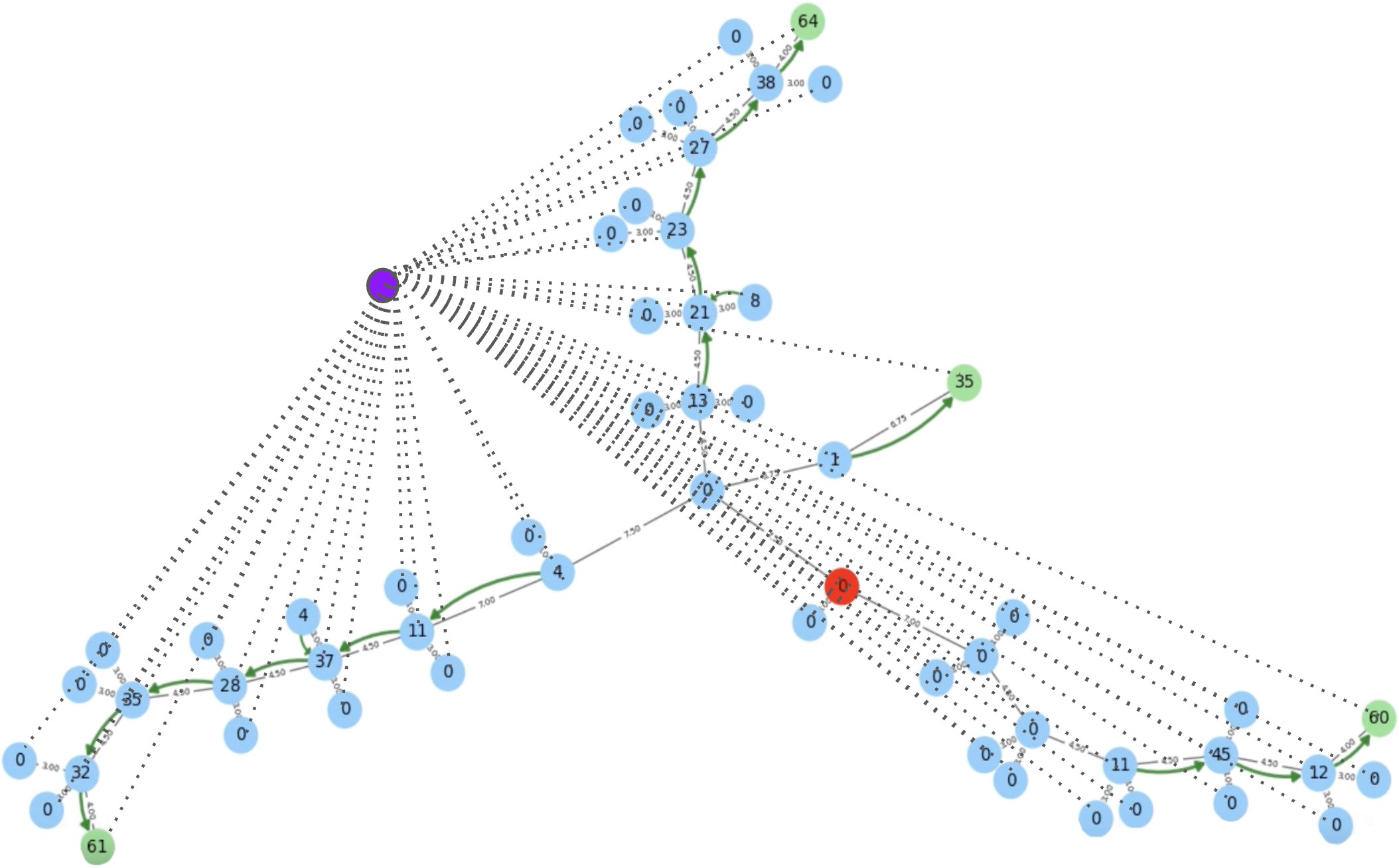}
    \caption{An illustration of the virtual global node (purple) and edges (dashed). The global node is connected to all other nodes (blue, green, and red).}
    \label{fig:global_node}
\end{figure}

As detailed in Section \ref{sec:method_gnn}, GPEVac incorporates a virtual global node connected to all other nodes within the sequential message-passing scheme. Both this global node and its corresponding edges are initialized as learnable parameters. This architecture addresses the limitations of standard GNNs on graphs with large diameters by enabling the network to efficiently capture long-range dependencies. A visualization of this global node is provided in Figure \ref{fig:global_node}.

\subsection{Masking}
\label{app:masking}

To allow a single policy to operate across varying layout sizes, observations are padded to fixed dimensional maxima for nodes, edges, and node degrees. To preserve the integrity of the graph topology, message passing operations explicitly ignore padded edges, and the virtual global node is connected exclusively to real nodes. Similarly, global pooling within the critic network masks out padded nodes to ensure that dummy slots do not corrupt the state-value estimate. 

Independent of this structural padding, GPEvac employs a dynamic action mask to enforce valid evacuation logic. At the node level, selection is restricted by masking out exit nodes, nodes devoid of evacuees, and nodes currently locked in a transit phase (i.e., $0<\text{time\_steps\_left}<T$). Furthermore, any topologically invalid routing actions from a given node are explicitly masked. Together, these structural and logical masks maintain a layout-agnostic architecture while ensuring that the learning gradient is updated solely by valid states and actions.

\subsection{Additional Implementation Details}
\label{app:implementation_details}

The GPEvac model evaluated in Table \ref{tab:evaluation_metrics} was trained for 18,000,000 environmental timesteps using a single NVIDIA Tesla V100 GPU. While the total wall-clock time for the complete training run was approximately 20 hours, the learning curve plateaued early; the network converged to its optimal routing policy within the first 8 hours of training. 


\section{Results}
\label{app:additional_results}

\subsection{Video Rollouts}
\label{app:video-rollout}
To best visualize GPEvac and the baselines in action, example rollout videos are provided in the 
supplementary material. For each architectural layout, we include a separate video demonstrating the performance of the GPEvac policy, the Rule-Based baseline, and the Greedy baseline. To ensure a direct and fair comparison, all three policies were evaluated using the exact same environment initialization seed.

\subsection{GPEvac}
\label{app:resuls-gpevac}

\begin{figure}[h]
\centering
\includegraphics[width=0.46\textwidth]{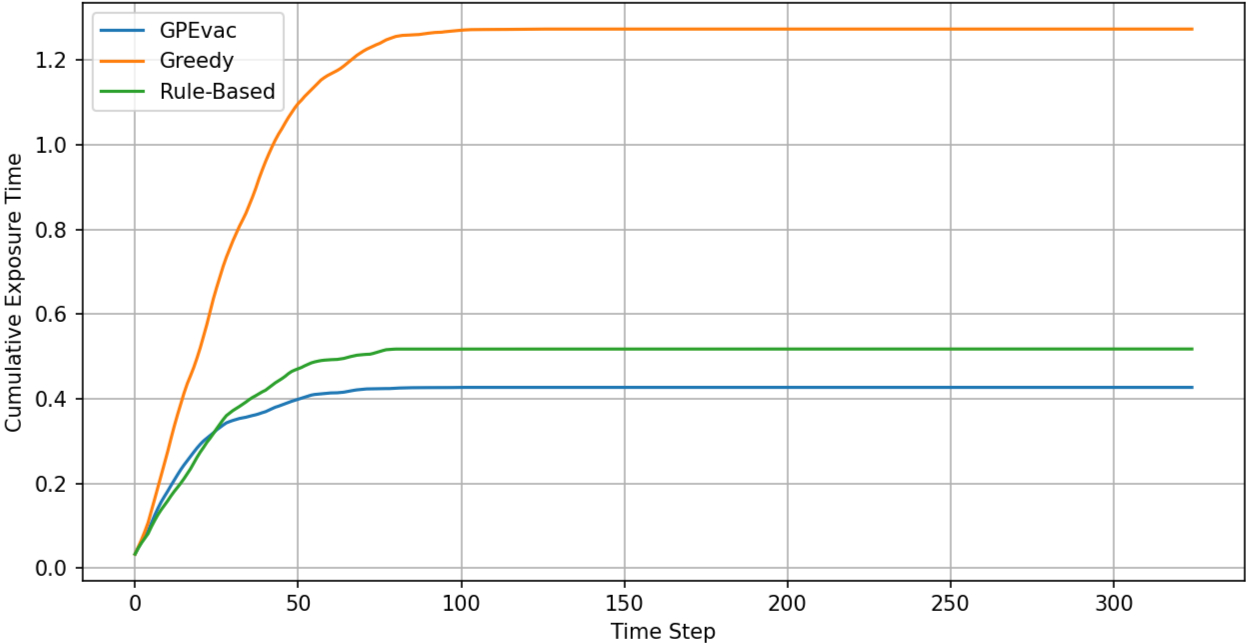}
\caption{The average cumulative exposure time of the best GPEvac policy at each time step. These are averaged across the test set of 32 environments per layout.}
\label{fig:results-through-stepping}
\end{figure}

Figure \ref{fig:results-through-stepping} details the step-wise performance of the optimal learned policy. It presents the average cumulative exposure time at each time step, aggregated across all evaluation episodes for both architectural layouts. GPEvac consistently outperforms both baselines in minimizing contact with the threat.

\subsection{Reward Function}
\label{app:rew_function}

\begin{table}[htbp]
\centering
\small
\setlength{\tabcolsep}{2.5pt} 
\begin{tabular}{@{} l|ccccc @{}}
    \toprule
    \textbf{Ratio} ($R_{\text{threat}}/R_{\text{exit}}$) & \textbf{0.1} & \textbf{0.25} & \textbf{1.0} & \textbf{2.0} & \textbf{5.0} \\
    \midrule
    \textbf{Threat Penalty} ($\downarrow$) & 5.800 & \textbf{4.430} & 9.646 & 7.494 & 36.758 \\
    \bottomrule
\end{tabular}
\caption{Normalized threat penalties of GPEvac across varying threat-to-exit reward ratios ($R_{\text{threat}}/R_{\text{exit}}$). Values represent the median threat penalty over 5 independent training runs on the validation set, normalized to match GPEvac's default weighting. At the default configuration ($R_{\text{threat}}/R_{\text{exit}} = 0.25$), GPEvac yields the lowest normalized threat penalty, outperforming the near-optimal Rule-Based baseline.}
\label{tab:reward_weight_ablation}
\end{table}

To evaluate the trade-off between the threat penalty and exit reward (Equations 1-5), we vary their ratio, $R_{threat}/R_{exit}$ (Table \ref{tab:reward_weight_ablation}). GPEvac consistently outperforms the Rule-Based baseline at $R_{threat}/R_{exit}=0.25$, which establishes an optimal equilibrium between speed and safety. However, deviating from this balance exposes two competing extremes: heavily prioritizing evacuation causes the policy to route evacuees directly into danger, whereas over-penalizing threat proximity instructs nodes to keep evacuees hidden, ultimately increasing overall exposure by delaying evacuation.

\subsection{Rule-Based Baseline}
\label{app:rule-based-baseline}

\begin{figure}[ht]
\centering
    \begin{minipage}[c]{0.43\textwidth}
    \centering
    \subfloat[Acyclic School\label{fig:rule_based_acyclic}]{\includegraphics[width=\textwidth]{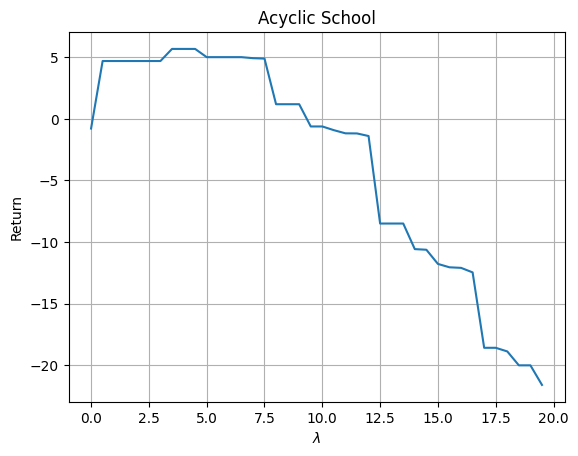}}
    \end{minipage}
    \hfill
    \begin{minipage}[c]{0.43\textwidth}
    \centering
    \subfloat[Cyclic School\label{fig:rule_based_cyclic}]{\includegraphics[width=\textwidth]{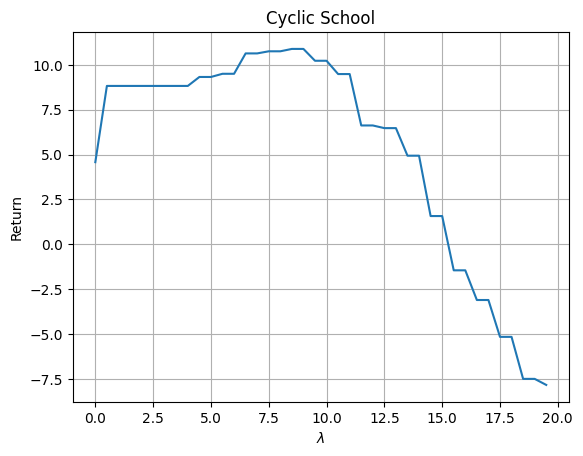}}
    \end{minipage}
\caption{The average return of the Rule-Based policy for each safety threshold $\lambda$. These values are averaged out across 100 random simulations of each layout.}
\label{fig:rule_based_grid_search}
\end{figure}

As explained in Section \ref{sec:baselines}, to determine the optimal safety thresholds ($\lambda$) for the Rule-Based baseline, we performed a grid search across a range of $\lambda$ values. For each school layout and $\lambda$ value, we evaluated the policy using 100 random simulations with identical seeds to ensure fair comparison (Figure~\ref{fig:rule_based_grid_search}). The Acyclic school achieved the highest average return at $\lambda = 4$, while the Cyclic school's return was maximized at $\lambda=9$.

\section{Broader Impacts and Limitations}
\label{sec:broader_impacts}

While the primary text addresses simulation and modeling constraints, deploying a real-world evacuation routing system introduces significant ethical and practical challenges. Future work must rigorously audit GPEvac to ensure it does not implicitly prioritize certain groups or building regions over others. Because GPEvac optimizes to minimize global contact with the threat, it risks learning utilitarian policies that could inadvertently sacrifice smaller groups of evacuees to protect the majority. Additionally, future iterations must explicitly explore how to best accommodate and protect vulnerable populations, such as the elderly, children, and individuals with disabilities \cite{wu2025toward}.

Furthermore, the psychological effects of active threat scenarios on evacuees pose a major modeling challenge~\cite{Mawson1978PanicBehavior}. Stress-induced behaviors, herd mentality, and panic are inherently difficult to simulate~\cite{Helbing2000SimulatingEscapePanic}. Consequently, future research must account for variable occupant compliance, specifically analyzing how highly stressed individuals are likely to perceive and react to dynamic routing directions provided via digital screens and smart exit signs.

Finally, the ``black-box'' nature of AI-based algorithms like GPEvac complicates their explainability. Although deep reinforcement learning often reduces threat contact more than mathematically rigorous, rule-based methods, the inability to pinpoint the exact reasoning behind specific routing decisions presents a major hurdle. Addressing this lack of interpretability will be critical for passing safety regulations, establishing trust, and securing legal certification for real-world deployment.

\end{document}